\PassOptionsToPackage{unicode}{hyperref}
\PassOptionsToPackage{hyphens}{url}
\PassOptionsToPackage{dvipsnames,svgnames,x11names}{xcolor}
\documentclass[12pt]{article}

\usepackage{amsmath,amssymb}
\usepackage{amsthm}
\usepackage{dsfont}
\usepackage{bbm}
\usepackage[shortlabels]{enumitem}
\usepackage{soul}

\usepackage{algorithm}
\usepackage{algpseudocode}

\usepackage{comment}

\newtheorem{theorem}{Theorem}[section] %
\newtheorem{lemma}[theorem]{Lemma}
\newtheorem{prop}[theorem]{Proposition}
\newtheorem{corollary}[theorem]{Corollary}

\DeclareMathOperator*{\argmax}{arg\,max}
\DeclareMathOperator*{\argmin}{arg\,min}

\usepackage[T1]{fontenc}
\usepackage[utf8]{inputenc}
\usepackage{textcomp}
\usepackage{lmodern}
\IfFileExists{upquote.sty}{\usepackage{upquote}}{}
\IfFileExists{microtype.sty}{%
  \usepackage[]{microtype}
  \UseMicrotypeSet[protrusion]{basicmath} %
}{}
\makeatletter
\@ifundefined{KOMAClassName}{%
  \IfFileExists{parskip.sty}{%
    \usepackage{parskip}
  }{%
    \setlength{\parindent}{0pt}
    \setlength{\parskip}{6pt plus 2pt minus 1pt}}
}{%
  \KOMAoptions{parskip=half}}
\makeatother
\usepackage{xcolor}
\makeatletter
\ifx\paragraph\undefined\else
  \let\oldparagraph\paragraph
  \renewcommand{\paragraph}{
    \@ifstar
      \xxxParagraphStar
      \xxxParagraphNoStar
  }
  \newcommand{\xxxParagraphStar}[1]{\oldparagraph*{#1}\mbox{}}
  \newcommand{\xxxParagraphNoStar}[1]{\oldparagraph{#1}\mbox{}}
\fi
\ifx\subparagraph\undefined\else
  \let\oldsubparagraph\subparagraph
  \renewcommand{\subparagraph}{
    \@ifstar
      \xxxSubParagraphStar
      \xxxSubParagraphNoStar
  }
  \newcommand{\xxxSubParagraphStar}[1]{\oldsubparagraph*{#1}\mbox{}}
  \newcommand{\xxxSubParagraphNoStar}[1]{\oldsubparagraph{#1}\mbox{}}
\fi
\makeatother

\usepackage{longtable,booktabs,array}
\usepackage{calc} %
\usepackage{etoolbox}
\makeatletter
\patchcmd\longtable{\par}{\if@noskipsec\mbox{}\fi\par}{}{}
\makeatother
\IfFileExists{footnotehyper.sty}{\usepackage{footnotehyper}}{\usepackage{footnote}}
\makesavenoteenv{longtable}
\usepackage{graphicx}
\makeatletter
\def\maxwidth{\ifdim\Gin@nat@width>\linewidth\linewidth\else\Gin@nat@width\fi}
\def\maxheight{\ifdim\Gin@nat@height>\textheight\textheight\else\Gin@nat@height\fi}
\makeatother
\setkeys{Gin}{width=\maxwidth,height=\maxheight,keepaspectratio}
\makeatletter
\def\fps@figure{htbp}
\makeatother

\makeatletter
\@ifpackageloaded{caption}{}{\usepackage{caption}}
\AtBeginDocument{%
\ifdefined\contentsname
  \renewcommand*\contentsname{Table of contents}
\else
  \newcommand\contentsname{Table of contents}
\fi
\ifdefined\listfigurename
  \renewcommand*\listfigurename{List of Figures}
\else
  \newcommand\listfigurename{List of Figures}
\fi
\ifdefined\listtablename
  \renewcommand*\listtablename{List of Tables}
\else
  \newcommand\listtablename{List of Tables}
\fi
\ifdefined\figurename
  \renewcommand*\figurename{Figure}
\else
  \newcommand\figurename{Figure}
\fi
\ifdefined\tablename
  \renewcommand*\tablename{Table}
\else
  \newcommand\tablename{Table}
\fi
}
\@ifpackageloaded{float}{}{\usepackage{float}}
\floatstyle{ruled}
\@ifundefined{c@chapter}{\newfloat{codelisting}{h}{lop}}{\newfloat{codelisting}{h}{lop}[chapter]}
\floatname{codelisting}{Listing}

\makeatother
\makeatletter
\@ifpackageloaded{caption}{}{\usepackage{caption}}
\@ifpackageloaded{subcaption}{}{\usepackage{subcaption}}
\makeatother

\usepackage[]{natbib}
\providecommand{\fullversion}{1}
\if0\fullversion
\usepackage{xr-hyper}
\makeatletter
\newcommand*{\addFileDependency}[1]{\typeout{(#1)}\@addtofilelist{#1}\IfFileExists{#1}{}{\typeout{No file #1.}}}
\makeatother
\newcommand*{\myexternaldocument}[1]{\externaldocument[][nocite]{#1}\addFileDependency{#1.tex}\addFileDependency{#1.aux}}
\makeatletter
\AtBeginDocument{\def\hyper@linkfile#1#2#3{#1}}%
\makeatother
\fi
\usepackage{bookmark}

\IfFileExists{xurl.sty}{\usepackage{xurl}}{} %
\hypersetup{
  pdftitle={Steering Bayesian Inference and Decisions through Data Poisoning},
  pdfauthor={Author 1; Author 2},
  pdfkeywords={3 to 6 keywords, that do not appear in the title},
  colorlinks=true,
  linkcolor={blue},
  filecolor={blue},
  citecolor={Blue},
  urlcolor={Blue},
  pdfcreator={LaTeX via pandoc}}
\if0\fullversion\myexternaldocument{supplement}\fi

\providecommand{\anon}{1}

\newcommand{\coderepo}{\if1\anon\url{https://github.com/roinaveiro/MMD_posterior_attraction}\else\url{https://anonymous.4open.science/r/MMD_posterior_attraction-EF14}\fi}

\begin{document}

\def\spacingset#1{\renewcommand{\baselinestretch}%
{#1}\small\normalsize} \spacingset{1}

\if1\anon
{
  \title{\bf Steering Bayesian Inference and Decisions\\ through Data Poisoning}
  \author{Roi Naveiro\thanks{
    Project carried out with the 2026 Leonardo Grant for Scientific Research and Cultural Creation from the BBVA Foundation. The BBVA Foundation accepts no responsibility for the opinions, comments, and content included in the project and/or the results derived from it, which are the sole and absolute responsibility of their authors. The views expressed in this article are those of the authors; they do not reflect the official policy or position of the U.S. Air Force or U.S. Government.}\hspace{.2cm}\\
    CUNEF Universidad, Madrid, Spain\\
    William N. Caballero \\
    Air Force Institute of Technology, OH, USA\\
    University of Arkansas, AR, USA\\
    and\\
    Juan Maroñas\thanks{Also with the Machine Learning Group, Universidad Autónoma de Madrid, Spain.}\\
    CUNEF Universidad, Madrid, Spain
    }
  \date{}
  \maketitle
} \fi

\if0\anon
{
  \bigskip
  \bigskip
  \bigskip
  \begin{center}
    {\LARGE\bf Steering Bayesian Inference and Decisions\\ through Data Poisoning}
\end{center}
  \medskip
} \fi

\bigskip
\begin{abstract}

Adversarial machine learning has revealed that statistical models can be manipulated through poisoned data, and Bayesian inference is no exception. Earlier work has shown that subtle poisoning attacks can drive the posterior toward a distribution of the attacker's choosing. Such posterior attraction attacks require a fully specified target posterior; however, attackers typically care about only a few posterior quantities or, more consequentially, about a downstream decision. Herein, we study how an attacker can alter Bayesian inference and decisions by deleting or replicating a few observations, chosen to minimize the maximum mean discrepancy between the resulting posterior and a target distribution. A suitable kernel lets the attacker steer selected posterior moments without specifying the rest of the target. Because Bayes actions depend on the posterior only through expected utilities, which are themselves posterior moments, the same approach can target a decision directly. We construct the attacks with a gradient-based heuristic that requires only posterior samples, and analyze its convergence. Experiments with regression and hierarchical models compare moment-based and full-posterior attacks, and illustrate how deleting a handful of observations can flip a Bayesian decision.

\end{abstract}

\noindent%
\vfill

\newpage
\spacingset{1.8} %

\newpage

\section{Introduction}\label{sec-intro}

Traditional Bayesian inference operates on a foundational but fragile assumption: data generation is a benign process. However, contemporary research in adversarial machine learning has shown how costly violations of this assumption can be. Modern statistical models must contend with environments where data are intentionally corrupted by self-interested agents who aim to manipulate model behavior.

Across many models from statistics' algorithmic culture \citep{breiman2001statistical}, such manipulation has been shown to severely degrade model performance. Once a model has been deployed, evasion attacks (that craft a data point's features but do not affect training data) have proven catastrophic to complex classifiers \citep{demetrio2021adversarial, chernikova2022fence, hong2022handcrafted}. They often entail perturbations that are imperceptible to humans \citep{goodfellow2014explaining, kurakin2018adversarial}. Alternatively, poisoning attacks modify training data during development, allowing the model's logic to be tailored to the attacker's desires. \citet{cina2023wild} provide a thorough survey of the data poisoning literature, covering the main variants and their effects. This line of research is more nascent in the data-modeling culture, especially with respect to Bayesian analysis. However, recent work by \citet{insua2023adversarial} and \citet{arce2025evasion} has highlighted its importance. Bayesian methods are fundamentally dependent on the integrity of the likelihood function. Its misspecification can substantively affect inference, a weakness that a strategically minded adversary can use to their advantage. Such is the focus of the posterior attraction attacks of \citet{pmlr-v258-carreau25a} and \citet{caballero2025posterior}. Given a fully specified target, i.e., \emph{adversarial posterior}, these works show how deleting, replicating or perturbing observations can let the attacker dictate inference.

In practice, however, an attacker rarely cares about every aspect of the posterior. Typically they want to shift a few posterior summaries, such as moments of the parameters or of downstream predictions. Requiring a full adversarial posterior commits them to details that are irrelevant to that goal, particularly in high-dimensional settings. We lift this requirement by replacing the Kullback--Leibler objective of \citet{pmlr-v258-carreau25a} with the maximum mean discrepancy \citep[MMD;][]{gretton2012kernel}. With a suitable kernel, the MMD compares only the summaries of interest, leaving the rest of the target unspecified. More consequentially, in most deployments the posterior is not an end in itself but an input to a downstream decision, and the attacker's ultimate aim is to change which action a decision maker adopts. This goal fits the same framework. Because a Bayes action depends on the posterior only through expected utilities (which are themselves posterior moments) inducing a target action reduces to driving the expected-utility gaps between competing actions across a decision boundary. As our experiments illustrate, Bayesian decisions can be surprisingly easy to flip.

To pursue these moment and decision targets, we manipulate the contribution of individual observations to the posterior. This connects our approach to the Bayesian coresets and case-deletion robustness literature. In this paper, following \citet{pmlr-v258-carreau25a}, an attack assigns each observation a nonnegative weight (0 for deleting, 1 for keeping, or greater than 1 for replicating) and thereby induces a \emph{weighted posterior}. This weighted posterior is precisely the object studied in Bayesian coreset constructions \citep{campbell2019automated, campbell2019sparse}, where the weights are chosen to compress a dataset while preserving its posterior; here they are instead chosen adversarially to steer it. The same object underlies the finite-sample robustness literature, which asks how few observations must be removed before a conclusion changes \citep{broderick2023automatic}. Whereas that literature evaluates the robustness of posterior inferences, we construct attacks. This connection is closest for the simplest, one-shot version of our attack, which uses a related first-order influence-ranking computation. We adopt the adversarial standpoint throughout, but the very same machinery can be read, in reverse, as a diagnostic for auditing the robustness of Bayesian decisions.

The remainder of this manuscript is structured as follows. Section \ref{sec:pa} formalizes the posterior attraction problem, reviewing the formulations of \citet{pmlr-v258-carreau25a} and \citet{caballero2025posterior} to keep the manuscript self-contained and to highlight their limitations. Section \ref{sec:mpa} sets forth an MMD-based version of the problem that can dispense with a full target posterior, derives a sampling-based gradient estimator, and presents special cases, including a reduction that lets the attacker target a downstream Bayes decision directly. Section \ref{sec:theory} provides theoretical analysis of the proposed attack. Sufficient conditions are derived to typify its performance and motivate the heuristics of Section \ref{sec:solution_methods}. Section \ref{sec:experiments} evaluates these heuristics on a simulated regression and two real datasets. Finally, Section \ref{sec-conc} provides closing remarks and discusses avenues for future inquiry. Proofs and additional details are provided in the \if0\fullversion Supplementary Material\else appendices\fi.

\section{Posterior Attraction Problems}\label{sec:pa}

Consider a decision maker (the \emph{Defender}, $D$) performing Bayesian inference on an unknown parameter $\theta \in \Theta \subseteq \mathbb{R}^d$. The agent observes a dataset $\mathbf{X} = (X_1,\ldots,X_n) \in \mathcal{X}^n$, where each observation $X_i$ takes values in a measurable space $(\mathcal{X},\mathcal{A})$.
Given a prior distribution $\pi(\theta)$ on $\Theta$ and likelihood functions $\pi(x \mid \theta)$ for $x \in \mathcal{X}$, the posterior distribution is
\begin{equation*}
\pi(\theta \mid \mathbf{X})
    = \frac{1}{Z(\mathbf{X})}
    \exp\!\left( \sum_{i=1}^n \log \pi(X_i \mid \theta) \right)
    \pi(\theta),
\end{equation*}
where $Z(\mathbf{X})$ is the marginal likelihood that ensures normalization.

For all but the simplest models, exact computation of $\pi(\theta \mid \mathbf{X})$ is intractable, so the Defender approximates it numerically, for example via MCMC or variational inference. Unless stated otherwise, we assume that $D$ uses MCMC. We write $\mathbb{E}_{\theta \sim P}[\cdot]$ for expectations with respect to $\theta \sim P$, dropping the subscript when it is clear from context.

We consider an adversarial setting where a second agent (the \emph{Attacker}, $A$) intercepts the uncorrupted dataset $\mathbf{X}$ and manipulates it to alter the Defender's inferences. The goal of such a manipulation is to steer the induced posterior as close as possible to a target posterior that encodes the Attacker's desired inferences. Formally, we study \emph{posterior attraction} problems \citep{pmlr-v258-carreau25a, caballero2025posterior} of the form
\begin{equation*}
\min_{\mathbf{X'} \in \mathcal{C}(\mathbf{X})}
 \delta\left[ \pi(\theta \mid \mathbf{X'}), \pi_A(\theta) \right],
\end{equation*}
where $\mathcal{C}(\mathbf{X})$ is the set of datasets that the Attacker can construct from $\mathbf{X}$ under the attack constraints, $\pi(\theta \mid \mathbf{X'})$ is the posterior induced by $\mathbf{X'}$, $\pi_A(\theta)$ is a target (adversarial) posterior distribution, and $\delta$ is a statistical discrepancy. We adopt a white-box setting, the natural benchmark for vulnerability to an informed attacker: $A$ knows the data, prior and likelihood and, when targeting a decision, $D$'s utility function.

Specific formulations of this problem are defined by two key choices: (i) the permissible data manipulations available to the attacker, and (ii) the statistical discrepancy $\delta$ used to measure success. Existing literature has explored the first axis while keeping the discrepancy fixed as the Kullback--Leibler (KL) divergence. The pioneering work of \citet{pmlr-v258-carreau25a} establishes the baseline model, adopting the KL divergence and restricting manipulations to the \emph{deletion} or \emph{replication} of existing points. Under this model, an attack is defined by a weight vector $w = (w_1, \ldots, w_n) \in \mathbb{Z}_{\ge 0}^n$, where $w_i = 0$ removes the $i^{th}$ observation, $w_i = 1$ leaves it unchanged, and $w_i = k > 1$ keeps $k$ copies of it. Let $f_{\mathbf{X}}(\theta) = \big( \log \pi(X_1 \mid \theta),\ldots, \log \pi(X_n \mid \theta) \big)^\top$ denote the vector of per-observation log-likelihood contributions. The posterior induced by an attack $w$ (the \emph{weighted posterior}) is then
\begin{equation} \label{eq:weighted_posterior}
\pi_w(\theta\mid \mathbf{X}) = \frac{1}{Z(w)} \exp \left(w^\top f_{\mathbf{X}}(\theta)\right)\pi(\theta),
\end{equation}
where $Z(w)$ is the marginal likelihood of the weighted dataset, assumed finite for every feasible $w$. To model limited resources and $A$'s desire to avoid detection, \citet{pmlr-v258-carreau25a} constrain the attack with a budget $B$ on the total number of deletions and additional copies and a limit $L$ on each observation's multiplicity. Taking the forward KL divergence as the objective gives the inclusive posterior attraction problem
\begin{equation} \label{eq:IPA}
    \min_{w \in \mathbb{Z}_{\ge 0}^n} \; \mathrm{KL}\left( \pi_A(\theta) \,\|\, \pi_w(\theta \mid \mathbf{X}) \right)
    \quad \text{s.t.} \quad \|w - \mathbf{1}\|_1 \le B, \;\; \|w\|_\infty \le L,
\end{equation}
\noindent where $\mathbf{1}$ denotes the length-$n$ vector of ones. They also study the exclusive variant, based on the reverse KL divergence. For most models, neither objective can be evaluated exactly, and the integer constraint on $w$ makes finding an optimal attack combinatorially hard. Addressing these difficulties, \citet{pmlr-v258-carreau25a} show that minimizing the KL divergence in \eqref{eq:IPA} is equivalent to minimizing $-w^\top \mathbb{E}_{\theta \sim \pi_A}[f_{\mathbf{X}}(\theta)] + \log Z(w)$, a convex function of $w$ with gradient $\mathbb{E}_{\theta \sim \pi_w}[f_{\mathbf{X}}(\theta)]-\mathbb{E}_{\theta \sim \pi_A}[f_{\mathbf{X}}(\theta)]$. This gradient can be estimated from samples of $\pi_w$ and $\pi_A$ (without bias if the draws are exact) even when the posterior is not analytically available. This suggests a powerful heuristic: relax the integer constraint, solve the continuous problem by projected stochastic gradient descent (SGD), and round the solution to an integer attack. \citet{pmlr-v258-carreau25a} provide convergence guarantees for the solution of the continuous relaxation through projected SGD and empirically demonstrate the effectiveness of this and related heuristics.

Seeking greater adversarial power, subsequent work by \citet{caballero2025posterior} extends this framework along the manipulation axis. Instead of deletions and replications, they allow the attacker to perturb the continuous values of the data $\mathbf{X}$. Their analysis covers exponential-family likelihoods with conjugate priors.

While the manipulation constraints differ, both works share a critical limitation: their reliance on the KL divergence, which forces the attacker to specify the \emph{entire} target posterior $\pi_A(\theta)$. In many practical scenarios, an attacker may only wish to alter specific features (such as posterior moments) rather than prescribing a fully specified distribution. In this paper, we explore the alternative axis of the design space. We retain the deletion/replication constraints but replace the KL divergence with the maximum mean discrepancy \citep[MMD;][]{gretton2012kernel}. As we shall see, this shift allows the attacker to express more general posterior-steering goals, such as those targeting specific moments, without needing to define a full adversarial posterior. More consequentially, it lets the attacker target the Defender's downstream decision directly. As we show in Section~\ref{sec:targeting_decisions}, inducing a prescribed Bayes action reduces to matching a set of expected-utility gaps.

\section{Posterior Attraction via MMD Minimization}\label{sec:mpa}

We begin by formally defining the MMD. Let $P$ and $Q$ be two probability distributions over the parameter space $\Theta$. Let $\mathcal{H}_k$ be a Reproducing Kernel Hilbert Space (RKHS) of functions $f: \Theta \to \mathbb{R}$, generated by a kernel $k: \Theta \times \Theta \to \mathbb{R}$. The MMD is defined as:
\begin{equation*}
    \text{MMD}(P, Q; k) \equiv \sup_{\substack{f \in \mathcal{H}_k \\ \|f\|_{\mathcal{H}_k} \le 1}} \left( \mathbb{E}_{\theta \sim P}[f(\theta)] - \mathbb{E}_{\theta' \sim Q}[f(\theta')] \right).
\end{equation*}
Intuitively, this optimization seeks a \textit{witness function} $f$ that maximizes the signal difference between $P$ and $Q$.  If the witness cannot find a distinction, the distributions are indistinguishable with respect to the features in $\mathcal{H}_k$. The reproducing property of the RKHS allows us to map each distribution to a \emph{kernel mean embedding} $\mu_P = \mathbb{E}_{\theta \sim P}[k(\theta, \cdot)] \in \mathcal{H}_k$. The MMD then corresponds to the distance between these embeddings, $\|\mu_P - \mu_Q\|_{\mathcal{H}_k}$. This duality allows the squared MMD to be expressed solely in terms of kernel evaluations, avoiding the intractable supremum \citep{gretton2012kernel}, i.e.,
\begin{equation} \label{eq:mmd_kernel}
    \text{MMD}^2(P, Q; k) = \mathbb{E}_{\theta, \theta' \sim P}[k(\theta, \theta')] - 2\mathbb{E}_{\theta \sim P, \theta' \sim Q}[k(\theta, \theta')] + \mathbb{E}_{\theta, \theta' \sim Q}[k(\theta, \theta')].
\end{equation}
The choice of kernel $k$ dictates the properties of the discrepancy. If $k$ is a \emph{characteristic kernel} (e.g., the RBF kernel), the MMD is a metric: $\text{MMD}(P, Q; k) = 0$ if and only if $P = Q$, so driving it to zero recovers the target distribution.

Using this framework, we formulate the \textit{MMD Posterior Attraction (MPA)} problem:
\begin{equation} \label{eq:MPA}
    \min_{w \in \mathbb{Z}_{\ge 0}^n} \; \mathrm{MMD}^2\left( \pi_A(\theta) , \pi_w(\theta \mid \mathbf{X}); ~k \right)
    \quad \text{s.t.} \quad \|w - \mathbf{1}\|_1 \le B, \;\; \|w\|_\infty \le L.
\end{equation}

A marginalization property makes precise that the attacker need not commit to a full adversarial posterior. Namely, split the parameters into $\theta=(\beta,\psi)$ where $\beta$ and $\psi$ are target and nuisance terms, respectively. If the kernel depends on $\theta$ only through $\beta$, i.e., $k(\theta,\theta')=k(\beta,\beta')$, the terms in equation \eqref{eq:mmd_kernel} depend on $\pi_A$ and $\pi_w$ only through their $\beta$-marginals, so the attacker need only prescribe a target for $\beta$. The KL objective of \citet{pmlr-v258-carreau25a} admits no such reduction: $\mathrm{KL}(\pi_A\,\|\,\pi_w)$ does not factor over $(\beta,\psi)$, so steering $\beta$ alone through it would still demand a fully specified joint target.

\subsection{Gradient Estimation via Sampling} \label{sec:mmd_min}

To address the intractability of the integer constraints in problem \eqref{eq:MPA}, we employ a relaxation-projection heuristic similar to that of \citet{pmlr-v258-carreau25a}. We relax the domain to continuous weights $w \in \mathbb{R}_{\ge 0}^n$, optimize the objective via projected SGD, and finally project the solution to the integer lattice. A key advantage of our optimization procedure is that it requires only the ability to sample from the posterior, without needing to evaluate the density or its gradient with respect to the parameters.

Performing SGD requires an unbiased estimator of the gradient of the squared MMD. We begin by establishing a general lemma, proved in Appendix~\ref{app:proof_grad_cov}, for differentiating expectations with respect to the weights $w$. This result relies on the structure of the weighted posterior $\pi_w(\theta \mid \mathbf{X})$ as an exponential family with natural parameters $w$ and sufficient statistics given by the log-likelihood vector $f_{\mathbf{X}}(\theta)$.

\begin{lemma}
\label{lemma:grad_cov}
Let $h: \Theta \to \mathbb{R}$ and assume that $|h(\theta)|$ and $\|f_{\mathbf{X}}(\theta)\|_2$ have finite second moments under $\pi_w$. If regularity conditions permit the interchange of differentiation and integration,
the gradient of the expectation $\mathbb{E}_{\theta \sim \pi_w}[h(\theta)]$ with respect to $w$ is
\begin{equation}
    \nabla_w \mathbb{E}_{\theta \sim \pi_w}[h(\theta)] = \mathrm{Cov}_{\pi_w}(h(\theta), f_{\mathbf{X}}(\theta)).
\end{equation}
\end{lemma}

The following theorem, proved in Appendix~\ref{app:proof_obj_grad}, derives the exact gradient of the squared MMD objective.

\begin{theorem} \label{thm:obj_grad}
Assume that $\|f_{\mathbf{X}}(\theta)\|_2$ has a finite second moment under $\pi_w$. Furthermore, assume the kernel satisfies $\mathbb{E}_{\theta \sim \pi_w}[k(\theta, \theta)] < \infty$ and $\mathbb{E}_{\theta \sim \pi_A}[k(\theta, \theta)] < \infty$, and that regularity conditions permit the interchange of integration and differentiation.

Define the function $g(\theta)$ as the difference between the kernel mean embeddings of the current posterior $\pi_w$ and the target $\pi_A$:
\begin{equation*}
    g(\theta) := \mathbb{E}_{\theta' \sim \pi_w}[k(\theta, \theta')] - \mathbb{E}_{\theta' \sim \pi_A}[k(\theta, \theta')].
\end{equation*}
The gradient of the squared MMD with respect to the weights $w$ is exactly twice the covariance between $g(\theta)$ and the vector of log-likelihoods $f_{\mathbf{X}}(\theta)$:
\begin{equation} \label{eq:true_gradient}
    \nabla_w \mathrm{MMD}^2(\pi_A, \pi_w; k) = 2 \cdot \mathrm{Cov}_{\pi_w}(g(\theta), f_{\mathbf{X}}(\theta)).
\end{equation}
\end{theorem}
This gradient admits an unbiased estimator even when the posterior $\pi_w$ is analytically intractable, provided we can sample from it. Algorithm \ref{alg:grad_est} details the procedure. Lemma~\ref{lemma:grad_unbiased} shows that this estimator is unbiased under exact sampling, as the convergence analysis of projected SGD requires (proof in Appendix~\ref{app:unbiased}).

\begin{algorithm}[htb]
\caption{Unbiased Gradient Estimator for MPA}
\label{alg:grad_est}
\begin{algorithmic}[1]
\State \textbf{Input:} Target samples $S_A \sim \pi_A$ (size $M$), Posterior samples $S_{\pi_w} \sim \pi_w$ (size $2N$).
\State \textbf{Split:} Partition $S_{\pi_w}$ into two disjoint sets $S_{\text{est}}$ and $S_{\text{cov}}$ of size $N$.
\State \textbf{Estimate Witness:} Construct $\hat{g}(\theta)$ using $S_{\text{est}}$ and $S_A$:
\begin{equation*}
    \hat{g}(\theta) \leftarrow \frac{1}{N} \sum_{\theta' \in S_{\text{est}}} k(\theta, \theta') - \frac{1}{M} \sum_{\theta' \in S_A} k(\theta, \theta')
\end{equation*}
\State \textbf{Compute Gradient:} Calculate sample covariance on held-out set $S_{\text{cov}}$, with $\bar{g}$ and $\bar{f}$ the means of $\hat{g}(\theta)$ and $f_{\mathbf{X}}(\theta)$ over $S_{\text{cov}}$:
\begin{equation*}
    \hat{\nabla} \leftarrow \frac{2}{N-1} \sum_{\theta \in S_{\text{cov}}} \left( \hat{g}(\theta) - \bar{g} \right) \left( f_{\mathbf{X}}(\theta) - \bar{f} \right)
\end{equation*}
\State \textbf{Return:} Unbiased gradient estimate $\hat{\nabla}$
\end{algorithmic}
\end{algorithm}

\subsection{Moment-matching Kernel}\label{sec:mm_kernel}

While characteristic kernels allow the attacker to target the full posterior, practical adversarial goals are often more specific. As discussed in Section~\ref{sec:pa}, an attacker may wish to manipulate only certain posterior summaries, such as posterior means. This objective can be naturally encoded within our MMD framework by selecting a kernel that explicitly measures the discrepancy of these summaries. Let $\{h_1, \dots, h_m\}$ be a set of functions $h_j: \Theta \to \mathbb{R}$ representing the features of interest (e.g., $h_1(\theta)=\beta_1$ targets the posterior mean of a regression coefficient $\beta_1$). To satisfy the integrability conditions of Theorem \ref{thm:obj_grad}, we assume that each $h_j(\theta)$ has a finite second moment under both the posterior $\pi_w$ and the target distribution $\pi_A$. We define the corresponding feature map $\phi(\theta) = [h_1(\theta), \dots, h_m(\theta)]^\top$ and construct the linear kernel $k_{\text{MM}}(\theta, \theta') = \phi(\theta)^\top \phi(\theta') = \sum_{j=1}^m h_j(\theta) h_j(\theta')$.
Substituting this kernel into the squared MMD definition \eqref{eq:mmd_kernel}, the objective function simplifies to the sum of squared errors between the expectations:
\begin{equation*}
    \mathrm{MMD}^2(\pi_A, \pi_w; k_{\text{MM}}) = \|\mathbb{E}_{\theta \sim \pi_A}[\phi(\theta)] - \mathbb{E}_{\theta \sim \pi_w}[\phi(\theta)]\|_2^2 = \sum_{j=1}^m \left( \mathbb{E}_{\theta \sim \pi_A}[h_j(\theta)] - \mathbb{E}_{\theta \sim \pi_w}[h_j(\theta)] \right)^2.
\end{equation*}
Thus, minimizing the MMD with this custom kernel is equivalent to matching the selected moments. Notice that $k_{\text{MM}}$ is generally not characteristic; the MMD can be zero even if $\pi_A(\theta) \neq \pi_w(\theta \mid  \mathbf{X} )$, provided their moments match. This lets the attacker meet specific goals without matching the target's full shape, which might require a much larger perturbation budget. We examine the trade-off between targeting specific summaries and reproducing the full posterior empirically in Section~\ref{sec:linreg}.

The objective function's gradient for this kernel assumes a particularly interpretable form. Let $\Delta_j = \mathbb{E}_{\theta' \sim \pi_w}[h_j(\theta')] - \mathbb{E}_{\theta' \sim \pi_A}[h_j(\theta')]$ be the error in the $j^{th}$ moment. The function $g(\theta)$ then becomes a weighted combination of the feature functions, $g(\theta) = \sum_{j=1}^m \Delta_j h_j(\theta)$.
Applying Theorem \ref{thm:obj_grad}, the gradient is the error-weighted sum of the covariances between the features and the likelihoods:
\begin{equation*}
    \nabla_w \mathrm{MMD}^2(\pi_A, \pi_w; k_{\text{MM}}) = 2 \sum_{j=1}^m \Delta_j \cdot \mathrm{Cov}_{\pi_w}(h_j(\theta), f_{\mathbf{X}}(\theta)).
\end{equation*}

For a given feature $h_j$ and observation $X_i$, its contribution to the gradient with respect to $w_i$ is proportional to the product of the discrepancy $\Delta_j$ with $\mathrm{Cov}_{\pi_w}(h_j(\theta), \log \pi(X_i \mid \theta))$. For instance, if the current posterior expectation of $h_j$ is higher than the target ($\Delta_j > 0$), an observation whose likelihood positively correlates with $h_j$ will produce a positive gradient. Gradient descent then decreases this observation's weight, penalizing data that pull the expected value in the wrong direction. Conversely, if the expectation is too low ($\Delta_j < 0$), the gradient becomes negative, naturally increasing the weight of that same helpful observation to push the mean upward. Computing this gradient requires only the targeted moments $\mathbb{E}_{\theta \sim \pi_A}[h_j(\theta)]$, not the full adversarial posterior $\pi_A$.

\subsection{Targeting Decisions} \label{sec:targeting_decisions}

The moment-matching kernel of the previous subsection lets an attacker steer chosen posterior summaries. Often, however, the attacker's real aim is to change which action the Defender takes. We now show that this goal also fits within the MPA framework: it reduces to matching a small set of posterior moments and thus inherits the gradient estimator of Section~\ref{sec:mmd_min}.

Consider first a finite action set $\mathcal{D}=\{d_1,\dots,d_K\}$ together with a utility function $u:\mathcal{D}\times\Theta\to\mathbb{R}$. Under the posterior $\pi_w(\theta\mid \mathbf{X})$, the Defender selects the Bayes action $d^\star(w)\in \argmax_{d\in\mathcal{D}} \mathbb{E}_{\theta\sim \pi_w}[u(d,\theta)]$.
Suppose the attacker wishes to induce a specific target action $d_A\in\mathcal{D}$. We introduce the utility gaps relative to $d_A$,
\begin{equation*}\label{eq:gap_features}
g_k(\theta)\;:=\;u(d_k,\theta)-u(d_A,\theta),\qquad k\in\{1,\dots,K\}\setminus\{A\}.
\end{equation*}
Then $d_A$ is the unique Bayes action precisely when every competitor yields strictly lower expected utility, $\mathbb{E}_{\theta\sim \pi_w}[g_k(\theta)] < 0$ for all $k\neq A$.
Inducing $d_A$ reduces to driving the $(K-1)$ posterior moments $\{\mathbb{E}_{\theta \sim \pi_w}[g_k]\}_{k\neq A}$ below zero, which we now encode through moment matching. We collect the gaps into the feature map $\phi(\theta)=(g_k(\theta))_{k\neq A}\in\mathbb{R}^{K-1}$ and form the associated linear kernel $k_{\mathrm{dec}}(\theta,\theta') := \phi(\theta)^\top \phi(\theta') = \sum_{k\neq A} g_k(\theta)\,g_k(\theta')$.
To satisfy the integrability conditions of Theorem \ref{thm:obj_grad}, we assume each gap $g_k(\theta)$ has a finite second moment under both $\pi_w$ and $\pi_A$. Encoding the attacker's intent as an adversarial moment specification $\mathbb{E}_{\theta\sim \pi_A}[\phi(\theta)] = m$ for some target $m\in\mathbb{R}^{K-1}$, the squared MMD under $k_{\mathrm{dec}}$ collapses, exactly as before, to $\mathrm{MMD}^2(\pi_A,\pi_w;k_{\mathrm{dec}}) = \| \mathbb{E}_{\theta\sim \pi_w}[\phi(\theta)] - m \|_2^2$.
Since the gaps must be strictly negative, we choose a negative target, fixing a margin $\gamma>0$ and setting $m=-\gamma\,\mathbf{1}\in\mathbb{R}^{K-1}$. Indeed, if an attack $w$ drives the objective below $\gamma^2$, that is $\mathrm{MMD}^2(\pi_A,\pi_w;k_{\mathrm{dec}}) < \gamma^2$, then $\|\mathbb{E}_{\theta\sim \pi_w}[\phi]-m\|_\infty \le \|\mathbb{E}_{\theta\sim \pi_w}[\phi]-m\|_2 < \gamma$, so each coordinate obeys
\begin{equation*}\label{eq:gap_negativity_from_mmd}
\mathbb{E}_{\theta\sim \pi_w}[g_k(\theta)]
\;=\; -\gamma + \Big(\mathbb{E}_{\theta\sim \pi_w}[\phi(\theta)]-m\Big)_k
\;<\;0,\qquad \forall k\neq A,
\end{equation*}
which is precisely the required negativity condition and hence guarantees $d^\star(w)=d_A$. The attacker can therefore target a decision without specifying a full target posterior: the attack is fully specified by the margin $\gamma$, and it suffices to solve the MPA problem \eqref{eq:MPA} with the decision kernel $k_{\mathrm{dec}}$ and $m=-\gamma\mathbf{1}$, estimating the gradient through Algorithm~\ref{alg:grad_est}. Because the negativity of the expected gaps is necessary and sufficient for $d_A$ to be the unique Bayes action, success can be certified, up to Monte Carlo precision, by estimating these gaps from posterior samples.

The same idea extends to continuous decisions, where the features are the gradients of the utility at the target action (Appendix~\ref{app:continuous}).

\section{Theoretical Analysis of Proposed Attack}\label{sec:theory}

In this section, we study performance guarantees for the proposed attacks, in which a candidate is obtained by solving the continuous relaxation of the MPA problem \eqref{eq:MPA} via projected SGD and subsequently mapped onto an integer feasible solution. These results inform the behavior of the attacks set forth in Section \ref{sec:solution_methods} and motivate the chosen heuristics. Throughout, we take $B \in \mathbb{Z}_{\ge 0}$ and $L \in \mathbb{Z}_{\ge 1}$ and denote the relaxed feasible region by $\mathcal{W} = \left \lbrace w \in \mathbb{R}^n \mid w \succeq 0, \ \|w\|_\infty \leq L, \ \| w - \mathbf{1} \|_1 \leq B \right \rbrace,$
which is non-empty, closed, bounded, and convex. We write $F(w) := \mathrm{MMD}^2(\pi_A, \pi_w; k)$ for the relaxed objective, and assume that the Euclidean projection $\Pi_\mathcal{W}(w) = \argmin_{w' \in \mathcal{W}} \|w' - w\|_2^2$ is available so that every iterate of projected SGD remains feasible. All results in this section are stated for a fixed observed dataset, and hence a fixed sample size $n$; the constants they involve, notably the smoothness modulus $\xi$, the initial gap $F(w_0)-F^\star$, and the variance bound $\sigma^2$, are problem-dependent and may themselves depend on $n$. We further assume $\mathcal{W}$ lies in the interior of the natural-parameter domain $\{w : Z(w) < \infty\}$, so that $\pi_w$ is proper for every $w \in \mathcal{W}$ and, by standard exponential-family arguments, the log-likelihoods $\log \pi(X_i \mid \theta)$ have finite moments of all orders under $\pi_w$, bounded uniformly over the compact set $\mathcal{W}$.

The form of the objective determines the kind of guarantee one can reasonably expect. Because the kernel mean embedding depends nonlinearly on the continuous variables $w$ through the weighted posterior $\pi_w$, the objective $F$ is in general nonconvex.
We therefore adopt the guarantee that is natural for constrained nonconvex stochastic optimization: projected SGD returns an approximate \emph{first-order stationary point} of the relaxation. When the objective does happen to be convex, this conclusion strengthens automatically, since stationarity then coincides with global optimality; we treat that situation in Appendix~\ref{app:convex}.

An appropriate optimality measure for a constrained optimization problem with a convex feasible region and a possibly nonconvex objective is defined via the gradient mapping
\begin{equation} \label{eq:pgm}
G_\eta(w) := \frac{1}{\eta}\left[ w - \Pi_\mathcal{W}\left\{ w - \eta \nabla F(w) \right\} \right], \qquad \eta > 0.
\end{equation}
That is, a point $w \in \mathcal{W}$ is first-order stationary for $\min_{w \in \mathcal{W}} F(w)$ if and only if $G_\eta(w) = 0$, equivalently if the inequality $\nabla F(w)^\top (z - w) \ge 0$ holds for all $z \in \mathcal{W}$, i.e., no feasible direction is one of first-order descent. When the projection is inactive, i.e., $w - \eta \nabla F(w) \in \mathcal{W}$, $G_\eta(w)$ reduces to $\nabla F(w)$, so $\|G_\eta(w)\|_2$ is the natural constrained analogue of the gradient norm, and it is the quantity that our convergence guarantee controls.

\subsection{Smoothness and the stochastic gradient oracle}

Our guarantee rests on three ingredients: curvature control through smoothness of the objective, unbiasedness of the gradient estimator $\hat{\nabla}$ produced by Algorithm \ref{alg:grad_est}, and a bound on its conditional variance. We establish each in turn.

The first ingredient controls how fast the gradient of $F$ can change across $\mathcal{W}$ (proof in Appendix~\ref{app:curvature}).

\begin{lemma} \label{lemma:Lsmooth}
Assume that $\mathbb{E}_{\theta \sim \pi_A}[k(\theta, \theta)] < \infty$, so that $\mu_A \in \mathcal{H}_k$, and that the kernel mean embedding $w \mapsto \mu_w$ is twice continuously Fr\'echet differentiable on an open neighborhood of $\mathcal{W}$. Then $\nabla F$ exists on $\mathcal{W}$, and the supremum of its Hessian operator norm,
\begin{equation*}
\xi := \sup_{w \in \mathcal{W}} \|\nabla^2_w F(w)\|_{\mathrm{op}} = \sup_{w \in \mathcal{W}} \left\| \, 2\langle \nabla_w \mu_w, \nabla_w \mu_w \rangle_{\mathcal{H}_k} - 2\langle \mu_A - \mu_w, \nabla^2_w \mu_w \rangle_{\mathcal{H}_k} \, \right\|_{\mathrm{op}},
\end{equation*}
is finite. Consequently, $F$ is $\xi$-smooth on $\mathcal{W}$ (i.e., $\|\nabla F(w) - \nabla F(v)\|_2 \le \xi \|w - v\|_2$), where $\mu_A$ and $\mu_w$ are the respective kernel mean embeddings of $\pi_A$ and $\pi_w$, and the inner products are taken component-wise.
\end{lemma}

The remaining two ingredients concern the stochastic gradient oracle, the routine that, given the current iterate $w$, returns a random estimate of $\nabla F(w)$. Here, the oracle is the estimator $\hat{\nabla}$ of Algorithm \ref{alg:grad_est}. When projected SGD is run, we write $\hat{\nabla}_t$ for the estimate this algorithm produces at iteration $t$. Letting $\mathcal{F}_t$ denote the filtration generated by the algorithm up to iteration $t$, Lemma \ref{lemma:grad_unbiased} (proved in Appendix~\ref{app:unbiased}) records the unbiasedness of this estimator; that is, conditioning on $w_t$, it yields $\mathbb{E}[\hat{\nabla}_t \mid \mathcal{F}_t] = \nabla F(w_t)$. This identity, and hence the guarantee below, presumes exact i.i.d.\ draws from $\pi_A$ and $\pi_w$; when the posterior is sampled by MCMC, as in our real-data experiments, the estimator need not be unbiased and the theorem does not directly apply without additional control of sampling bias and dependence.

\begin{lemma} \label{lemma:grad_unbiased}
Assume the conditions of Theorem \ref{thm:obj_grad} hold and that the batches $S_A$, $S_{\text{est}}$ and $S_{\text{cov}}$ are mutually independent and contain exact i.i.d.\ draws from their respective distributions, with $M \ge 1$ and $N \ge 2$. Then $\hat{\nabla}$ (Algorithm \ref{alg:grad_est}) is an unbiased estimator of $\nabla_w \mathrm{MMD}^2(\pi_A, \pi_w; k)$.
\end{lemma}

The third ingredient bounds the dispersion of this estimator around its mean. Because $\hat{\nabla}_t$ is unbiased, its conditional variance is at most its conditional second moment, which the next lemma bounds uniformly over $\mathcal{W}$ under a fourth-moment condition on the feature norm $\|k(\theta,\cdot)\|_{\mathcal{H}_k} = \sqrt{k(\theta,\theta)}$ (proof in Appendix~\ref{app:variance}).

\begin{lemma} \label{lemma:gradient_normsq}
Under the assumptions of Lemma~\ref{lemma:grad_unbiased}, if $\sup_{w \in \mathcal{W}} \mathbb{E}_{\theta \sim \pi_w}[k(\theta,\theta)^2] < \infty$ and $\mathbb{E}_{\theta \sim \pi_A}[k(\theta,\theta)^2] < \infty$, then there exists $\sigma^2 < \infty$ such that the estimator of Algorithm \ref{alg:grad_est} obeys $\mathbb{E}[\|\hat{\nabla}_t\|_2^2 \mid \mathcal{F}_t] \le \sigma^2$, and hence $\mathbb{E}[\|\hat{\nabla}_t - \nabla F(w_t)\|_2^2 \mid \mathcal{F}_t] \le \sigma^2$, for all $w_t \in \mathcal{W}$.
\end{lemma}

\subsection{Convergence to a stationary point}

We can now state the central guarantee, which adapts the analysis of \citet{ghadimi2016mini} (proof in Appendix~\ref{app:psgd}). The result assures that, if we terminate the algorithm at iteration $R$ chosen uniformly at random from $\{0,\cdots,T-1\}$, the expected squared norm of the gradient mapping is bounded by a term that decays as $\mathcal{O}(1/T)$, contingent upon the starting position, plus a term driven by the stochastic gradient's variance.

\begin{theorem} \label{thm:psgd_stationary}
Suppose $F$ is $\xi$-smooth on $\mathcal{W}$ for some $\xi > 0$ and that the gradient estimator $\hat{\nabla}_t$ is conditionally unbiased, $\mathbb{E}[\hat{\nabla}_t \mid \mathcal{F}_t] = \nabla F(w_t)$ (Lemma \ref{lemma:grad_unbiased}), with a finite deterministic conditional-variance bound $\nu_t^2$ at each iteration,
\begin{equation*}
\mathbb{E}\!\left[ \|\hat{\nabla}_t - \nabla F(w_t)\|_2^2 \mid \mathcal{F}_t \right] \le \nu_t^2 ;
\end{equation*}
Lemma \ref{lemma:gradient_normsq} guarantees that such a bound exists. Run projected SGD from a fixed $w_0 \in \mathcal{W}$ with constant step size $\eta = 1/(2\xi)$,
\begin{equation*}
w_{t+1} = \Pi_\mathcal{W}\left( w_t - \eta \, \hat{\nabla}_t \right), \qquad t = 0,\dots,T-1,
\end{equation*}
and let $R$ be drawn uniformly from $\{0,\dots,T-1\}$, independently of the algorithm. Writing $F^\star := \inf_{w \in \mathcal{W}} F(w)$, the gradient mapping \eqref{eq:pgm} satisfies
\begin{equation} \label{eq:stationarity_bound}
\mathbb{E}\left[ \|G_\eta(w_R)\|_2^2 \right] \le \frac{8\xi\left( F(w_0) - F^\star \right)}{T} + \frac{6}{T} \sum_{t=0}^{T-1} \nu_t^2.
\end{equation}
\end{theorem}

With a fixed number of samples per iteration, Lemma~\ref{lemma:gradient_normsq} allows $\nu_t^2 = \sigma^2$, and the bound becomes $8\xi(F(w_0) - F^\star)/T + 6\sigma^2$, which does not vanish as $T$ grows: the theorem then guarantees only a stationarity residual $\mathbb{E}\|G_\eta(w_R)\|_2$ of order $\sigma$. Making the bound vanish requires $T^{-1}\sum_t \nu_t^2 \to 0$, for instance through growing per-iteration sample sizes with a variance bound that shrinks with them, a refinement we do not develop. Two further caveats apply: the step size $\eta = 1/(2\xi)$ involves the unknown smoothness modulus, so in practice step sizes are tuned empirically; and the guarantee concerns a uniformly sampled iterate $w_R$, whereas Algorithm~\ref{alg:sgd_r2} rounds the last iterate.

\subsection{Integer attacks}

The attacks of Section \ref{sec:solution_methods} return an integer feasible solution by rounding the relaxed iterate $w_T$ onto $\mathbb{Z}^n_{\ge 0} \cap \mathcal{W}$. We write this rounded image as $\hat{w} := \Pi_{\mathbb{Z}^n_{\ge 0} \cap \mathcal{W}}(w_T)$, the integer feasible point closest to $w_T$ in Euclidean distance, which the constrained rounding of Proposition \ref{prop:rounding} computes exactly. Smoothness lets us bound the objective distortion introduced by this step. By Lemma~\ref{lemma:Lsmooth} and the convexity of $\mathcal{W}$, we have
\begin{equation} \label{eq:rounding_distortion}
F(\hat{w}) - F(w_T) \le \nabla F(w_T)^\top \left( \hat{w} - w_T \right) + \frac{\xi}{2} \, \| \hat{w} - w_T \|_2^2.
\end{equation}
This bound quantifies the price of integrality relative to the point actually reached, which is small whenever $w_T$ lies close to the integer lattice.
Appendix~\ref{app:convex} sharpens the bound in the convex case, where the distortion can be measured against the integer optimum itself.

\section{Solution Methods} \label{sec:solution_methods}

Since MPA problem \eqref{eq:MPA} optimizes over integer weights, gradient methods cannot be applied directly. We propose two families of attacks that nonetheless exploit the gradient estimator of Algorithm~\ref{alg:grad_est}: a continuous relaxation, solved by projected SGD and then rounded, and faster greedy heuristics that take integer steps guided by the estimated gradient.

\subsection{Rounded Relaxation (SGD-R2)}

Our primary method, SGD Rounded Relaxation (SGD-R2), implements the strategy motivated in Section \ref{sec:mmd_min} in two phases. We first relax the integer constraint $w \in \mathbb{Z}_{\ge 0}^n$ to $w \in \mathbb{R}^n$ and optimize over the relaxed feasible set $\mathcal{W}$ by projected SGD. At each iteration $t$, we draw samples from the current posterior $\pi_{w_t}$ and the target $\pi_A$. Thereafter, we form the unbiased gradient estimate $\hat{\nabla}_t$ with Algorithm~\ref{alg:grad_est}, and update the weights, projecting back onto $\mathcal{W}$ through the Euclidean projection $\Pi_\mathcal{W}$, which, since $\mathcal{W}$ is a box intersected with an $\ell_1$-ball, can be computed in $\mathcal{O}(n\log n)$ time. The final iterate $w_T$ is generally non-integer, so we then round it to the closest feasible integer point, $\hat{w} = \Pi_{\mathbb{Z}^n_{\geq 0} \cap \mathcal{W}}(w_T)$, using the following closed form, established by \citet{pmlr-v258-carreau25a} for the same feasible set.

\begin{prop} \label{prop:rounding}
For $v \in \mathcal{W}$, let $\Delta_i = |v_i - 1|$ and $N_{\text{max}} = B - \sum_i \lfloor \Delta_i \rfloor$. A closest point to $v$ in $\mathbb{Z}^n_{\geq 0} \cap \mathcal{W}$ has coordinates $1 + \operatorname{sign}(v_i - 1)(\lfloor \Delta_i \rfloor + \alpha_i)$, where $\alpha_i = 1$ for every $i$ whose fractional part $\Delta_i - \lfloor \Delta_i \rfloor$ exceeds $1/2$, if there are at most $N_{\text{max}}$ such indices, and otherwise for the $N_{\text{max}}$ indices with the largest fractional parts; all other $\alpha_i$ equal $0$.
\end{prop}

Algorithm~\ref{alg:sgd_r2} summarizes the procedure. Because $\pi_{w_t}$ changes with every update, fresh samples are needed at each iteration and, with MCMC, this resampling is the main computational cost. We reduce it by \emph{warm-starting}: the chains at iteration $t+1$ start from the final states of iteration $t$, which shortens burn-in. MCMC draws are correlated and only approximately distributed as $\pi_{w_t}$, and warm-starting also makes them depend on earlier iterations, so the unbiasedness behind Section~\ref{sec:theory} holds only approximately. Our experiments (Section~\ref{sec:experiments}) nevertheless show high-quality attacks. Replacing the SGD step by an Adam step \citep{kingma2015adam} before the projection gives a faster variant, Adam-R2.

\begin{algorithm}[htb]
\caption{SGD Rounded Relaxation (SGD-R2)}\label{alg:sgd_r2}
\begin{algorithmic}[1]
\State \textbf{Input:} Initial weight $w_0 \in \mathcal{W}$ (e.g., $w_0=\mathbf{1}$), learning rate schedule $(\eta_t)_{t=0}^\infty$, sample sizes $M$ and $N$; set $t = 0$.
\Repeat
    \State Sample $S_A \sim \pi_A$ (size $M$) and $S_{\pi_{w_t}} \sim \pi_{w_t}$ (size $2N$).
    \State Estimate the gradient $\hat{\nabla}_t$ with Algorithm~\ref{alg:grad_est} from $S_A$ and $S_{\pi_{w_t}}$.
    \State Update $w_{t+1} \gets \Pi_\mathcal{W}(w_t - \eta_t \hat{\nabla}_t)$ and $t \gets t + 1$.
\Until{stopping criterion is met}
\State \Return $\hat{w} = \Pi_{\mathbb{Z}^n_{\geq 0} \cap \mathcal{W}}(w_T)$, with $T$ the final iteration, computed by Proposition~\ref{prop:rounding}.
\end{algorithmic}
\end{algorithm}

\subsection{Other heuristics}

SGD-R2 can be computationally expensive, since the repeated posterior sampling required for convergence is costly, and the optimum of the continuous relaxation need not lie near the integer optimum. We therefore propose two lighter-weight heuristics that maintain integrality throughout the search, dispensing with the rounding step.

\paragraph*{Fast Gradient Sign Method (FGSM).} This is a one-shot attack that serves as a baseline. We compute a single gradient estimate $\hat{\nabla}$ at the initial state $w=\mathbf{1}$ using Algorithm \ref{alg:grad_est}, select the $B$ observations with the largest $|\hat{\nabla}_i|$, and move each one unit against the sign of its gradient, deleting it if $\hat{\nabla}_i > 0$ and adding a copy if $\hat{\nabla}_i < 0$. The deletion-only variant deletes only observations with $\hat{\nabla}_i > 0$, so it may use less than the full budget.

This one-shot construction links our attacks to the finite-sample robustness literature, which gauges how a posterior conclusion would change were a few observations removed. The Approximate Maximum Influence Perturbation of \citet{broderick2023automatic}, together with its MCMC extension by \citet{nguyen2024sensitivity}, ranks observations by a single influence score computed once at the clean posterior; FGSM performs the same one-shot ranking for the MMD objective. When the attack is restricted to deletions and targets a single posterior moment at a value different from its clean value, the two rankings coincide once the influence diagnostic is aimed in the direction of the target, so deletion-only FGSM is essentially an instance of this influence diagnostic. The iterative heuristic below departs from it by recomputing the ranking as the attacked posterior evolves.

\paragraph*{Integer-Steps Coordinate Descent (ISCD).} This method refines FGSM by applying perturbations iteratively, one unit at a time. At each step, it re-estimates the gradient at the current $w$ and, among the feasible single-coordinate moves, takes the one with the largest first-order decrease of the objective,
\begin{equation*}
j \gets \argmax_{i \in \{1,\dots,n\}} \left\{ |\hat{\nabla}_i| \;\middle|\; w - \operatorname{sign}(\hat{\nabla}_i)e_i \in \mathcal{W} \right\}, \qquad w \gets w - \operatorname{sign}(\hat{\nabla}_j)e_j,
\end{equation*}
where $e_i$ is the $i$th canonical basis vector, until the budget is exhausted or a maximum number of iterations is reached (Algorithm~\ref{alg:iscd}, Appendix~\ref{app:algorithms}).

\section{Empirical Evaluation}
\label{sec:experiments}

To assess the effectiveness of our proposed MPA framework, we conduct a series of targeted data poisoning experiments.\footnote{Code and data to reproduce all experiments are available at \coderepo.}

\subsection{Bayesian Linear Regression with Unknown Variance}
\label{sec:linreg}

This subsection uses a controlled simulation study to examine two design choices in the proposed framework: (1) the choice of MMD kernel and (2) the choice of attack heuristic. We also illustrate how the advantage of moment matching over full-posterior matching depends on the geometry of the posterior. The Defender fits a Bayesian linear regression model in which both the regression coefficients and the observation variance are unknown, with a conjugate Normal-Inverse-Gamma prior:%
\begin{align*}
    y_i \mid x_i, \beta, \sigma^2 &\sim \mathcal{N}(x_i^\top \beta, \sigma^2), \quad i = 1, \dots, n, \\
    \beta \mid \sigma^2 &\sim \mathcal{N}(\mu_0, \sigma^2 \Sigma_0), \\
    \sigma^{-2} &\sim \mathrm{Gamma}(a_0, b_0).
\end{align*}
The model parameter is thus $\theta=(\beta,\sigma^2)$.
For any weight vector $w \in \mathbb{R}_{\ge 0}^n$, conjugacy implies that the weighted posterior remains Normal-Inverse-Gamma. Its parameters are derived in Appendix~\ref{app:nig}.
Exact posterior draws are therefore available under every attack, so the MMD objective and its stochastic gradients can be evaluated without introducing MCMC approximation error. Appendix~\ref{app:grad_check} uses this setting to check Lemma~\ref{lemma:grad_unbiased} numerically.

We generate a dataset with $n = 200$ observations and $d = 2$ covariates, without an intercept. The design vectors are sampled from a centered bivariate normal distribution with correlation $\rho = 0.8$, the data-generating coefficient is $\beta = (\beta_1,\beta_2)^\top = (2.0, -1.0)^\top$, and the noise standard deviation is $\sigma = 1.0$. We use the weakly informative prior $\mu_0 = (0,0)^\top$, $\Sigma_0 = 25^2 I_2$, and $a_0 = b_0 = 10^{-3}$.%

The attacker seeks to shift the posterior mean of the first regression coefficient, $\beta_1$, from its clean value, which is close to $2$, to the target value $0$.
We compare two kernels. Under the moment-matching kernel, the attack targets the single feature $h_1(\theta)=\beta_1$.
 The target value for the moment is set to $\mathbb{E}_{\theta \sim \pi_A}[\beta_1] = 0$. Under the RBF kernel, the adversarial posterior is constructed from the clean posterior itself. Specifically, if $(\mu_n,\Sigma_n,a_n,b_n)$ denote the parameters of the Normal-Inverse-Gamma clean posterior, the adversarial posterior is chosen to be a Normal-Inverse-Gamma distribution with parameters $(\tilde{\mu}_n,\Sigma_n,a_n,b_n)$, where $\tilde{\mu}_n$ is the same as $\mu_n$ except that its first component is set to $0$. For the RBF kernel bandwidth, we use the median heuristic~\citep{gretton2012kernel, garreau2017large}, computed once on a pooled pilot sample from the clean and adversarial posteriors and kept fixed during the attack.

We compare attacks found by four heuristics (i.e., FGSM, ISCD,  SGD-R2, and Adam-R2; Section~\ref{sec:solution_methods} and Appendix~\ref{app:algorithms}) using both kernels, over budgets $B \in \{0,10,20,\dots,100\}$ (up to half of the sample; deliberately large, for illustration), with maximum per-point replication limit of $L=2$. To account for the stochasticity in the gradient estimates, we repeat each attack 10 times and report the average plus/minus two standard deviations. Further implementation details are given in Appendix~\ref{app:settings}.

\begin{figure}[tb]
    \centering
    \begin{subfigure}[t]{0.46\linewidth}
        \centering
        \includegraphics[width=\linewidth]{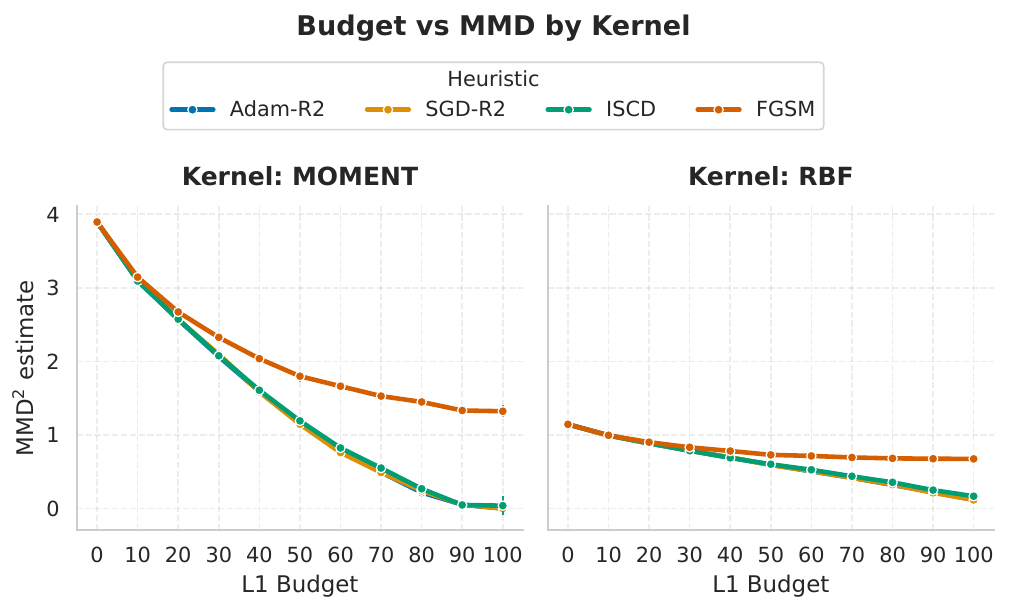}
        \caption{Achieved squared-MMD objective across heuristics.}
        \label{fig:linreg_unknown_var_mmd_comparison}
    \end{subfigure}\hfill
    \begin{subfigure}[t]{0.46\linewidth}
        \centering
        \includegraphics[width=\linewidth]{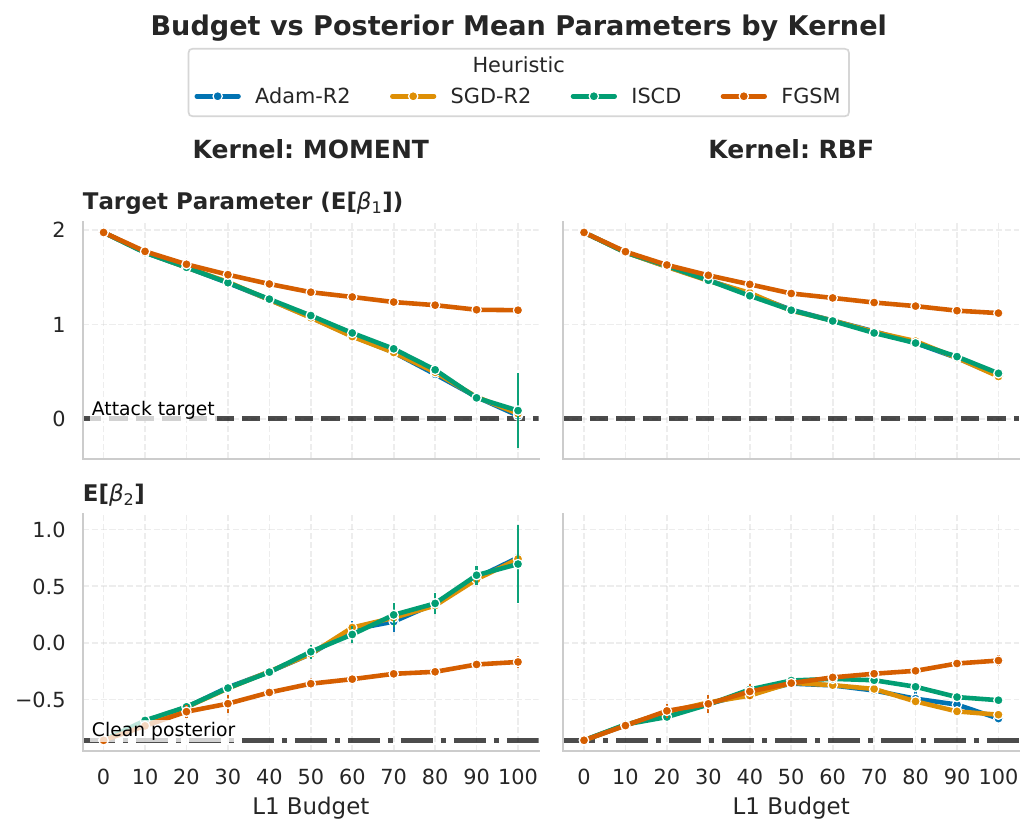}
        \caption{Induced posterior means $\mathbb{E}_{\theta \sim \pi_w}[\beta_1]$ (targeted) and $\mathbb{E}_{\theta \sim \pi_w}[\beta_2]$ (non-targeted).}
        \label{fig:linreg_unknown_var_posterior_parameters}
    \end{subfigure}
    \caption{Linear-regression attack as a function of the perturbation budget $B$, by heuristic and kernel.}
    \label{fig:linreg_unknown_var_budget}
\end{figure}

Figure~\ref{fig:linreg_unknown_var_budget} illustrates how the attacks evolve with the budget $B$: panel~(a) shows the attained MMD$^2$, and panel~(b) the induced posterior means of the targeted coefficient $\beta_1$ and the untargeted coefficient $\beta_2$. Within each kernel, the three iterative heuristics Adam-R2, SGD-R2, and ISCD behave similarly in terms of the attained MMD objective, whereas FGSM consistently plateaus at substantially larger discrepancy values. The main difference among the iterative methods is stability: the attack found by ISCD can be more variable for large budgets. FGSM plateaus because it ranks the observations only once, by their influence on the clean posterior; as the attack moves the posterior, this ranking becomes stale and the changes made at larger budgets contribute little. The iterative heuristics avoid this by re-estimating the gradient at the current weights, at the price of many gradient estimates instead of one.

\begin{figure}[tb]
    \centering
    \includegraphics[width=0.72\linewidth]{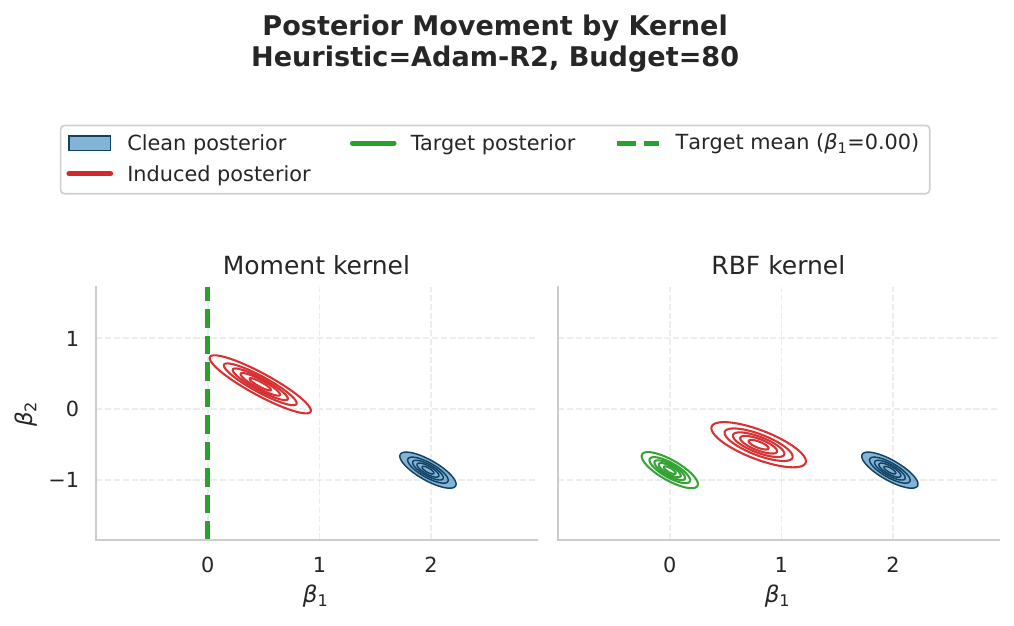}
    \caption{Posterior movement at budget $B=80$ for Adam-R2: clean, attacked, and target posteriors of $(\beta_1,\beta_2)$ under the moment-matching (left) and RBF (right) kernels. For each kernel, the run with the smallest MMD$^2$ is shown.}
    \label{fig:linreg_unknown_var_posterior_movement}
\end{figure}

The substantive distinction is between kernels. Under the moment-matching kernel, the attack is relatively more effective. By budget $B=100$%
, the induced posterior mean of $\beta_1$ is driven essentially to $0$ by the three iterative heuristics.
This success comes with collateral distortion in the untargeted coefficient $\beta_2$, whose posterior mean moves from its clean value near $-0.8$ to approximately $0.7$.
Under the RBF kernel, by contrast, the attack produces a much smaller displacement in $\mathbb{E}_{\theta \sim \pi_w}[\beta_1]$, which falls only to about $0.5$ by budget $B=100$ for the iterative heuristics, but it also preserves the untargeted posterior mean much more closely, with $\mathbb{E}_{\theta \sim \pi_w}[\beta_2]$ ending near $-0.6$.  Figure~\ref{fig:linreg_unknown_var_posterior_movement} illustrates where this gap comes from, showing the clean, attacked, and target posteriors at $B=80$. The gap is driven by what each objective constrains. The moment kernel constrains only $\mathbb{E}_{\theta \sim \pi_w}[\beta_1]$, so the attack moves the posterior along the ridge induced by the correlated design, dragging $\beta_2$ with it. The RBF target also fixes the mean of $\beta_2$ and the spread and correlation of the posterior at their clean values. This target lies off the ridge, so reweighting reaches it only partially, but the attack keeps $\beta_2$ closer to its clean value. Both attacks also widen the posterior. Moment matching thus suits an attacker who cares about one functional, and the RBF kernel one who wants the remaining features held at specified values.

\begin{figure}[tb]
    \centering
    \includegraphics[width=0.9\linewidth]{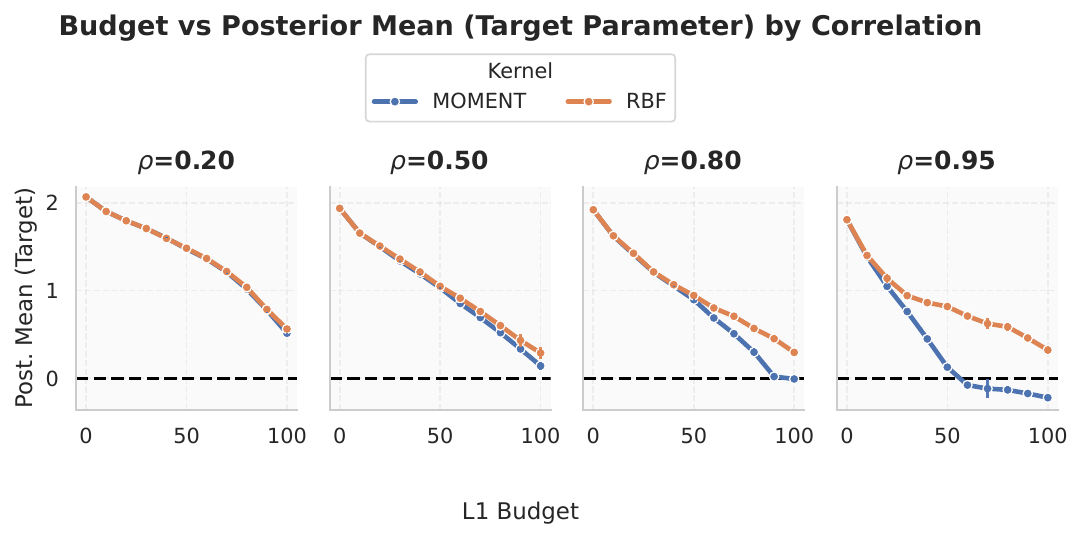}
    \caption{
    Sensitivity of the targeted posterior mean, under the Adam-R2 attack, to the population covariate correlation $\rho$ of the data-generating process, with unit covariate variances and one simulated dataset per $\rho$. The dashed line marks the target value $0$.
    }
    \label{fig:linreg_unknown_var_rho_sensitivity}
\end{figure}

Figure~\ref{fig:linreg_unknown_var_rho_sensitivity} illustrates how this gap depends on the covariate correlation $\rho$. The two kernels coincide at $\rho=0.2$, and the gap widens as higher correlation sharpens the posterior ridge. At $\rho=0.95$ and large budgets the moment attack overshoots the target: the relaxed iterate sits at the target, and the overshoot comes from the final rounding step.

These attacks are not limited to posterior means: they can also target uncertainty estimates. Appendix~\ref{exp:attack_var} repeats the experiment with the posterior variance of $\beta_1$ as the target, which the moment-matching attack inflates by more than an order of magnitude while keeping the posterior mean close to its clean value, so that credible intervals widen while the point estimate barely moves.

\subsection{Bayesian Spatial Regression}
\label{sec:spatial_regression}

\begin{figure}[tb]
    \centering
    \includegraphics[width=\linewidth]{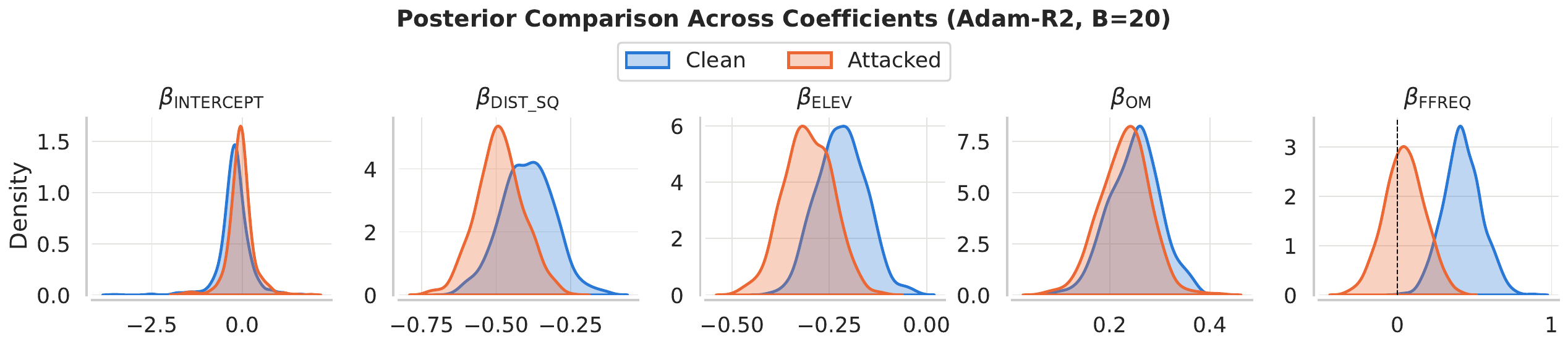}
    \caption{Posterior comparison across regression coefficients for the selected budget-$20$ attack (Adam-R2).}
    \label{fig:spatial_posteriors}
\end{figure}

We now turn to a setting in which the posterior is not available in closed form, requiring MCMC-based inference. We
use the \textit{Meuse} soil survey \citep{rikken1993soil}, comprising $N=152$ stream-bank soil samples collected along the Meuse river. Each sample is spatially referenced by its map coordinates (in kilometers) and records the concentration of heavy metals in the floodplain topsoil.
The response variable is the standardized log-zinc concentration. We include four covariates: the square root of the distance to the riverbank ($\mathrm{DIST\_SQ}$), standardized elevation ($\mathrm{ELEV}$), standardized organic-matter content ($\mathrm{OM}$), and a flooding-frequency indicator ($\mathrm{FFREQ}$), equal to $1$ for sites flooded every 1--2 years and $0$ for sites flooded less often. Suppose
a local regulator models the response $y_i$ at location $s_i \in \mathbb{R}^2$ using a non-conjugate spatial regression:
$$
y_i = x_i^\top \beta + u(s_i) + \varepsilon_i, \qquad \varepsilon_i \sim \mathcal{N}(0,\sigma_\varepsilon^2),
$$
where the latent spatial effect $u$ is a zero-mean Gaussian process with covariance function $\operatorname{Cov}(u(s_i),u(s_j)) = \sigma_u^2 \exp(-\phi\|s_i-s_j\|_2)$.
The regulator places weakly informative priors $\beta_j \sim \mathcal{N}(0,10^2)$, $\sigma_\varepsilon,\sigma_u \sim \mathrm{HalfCauchy}(5)$, and $\phi \sim \mathrm{LogNormal}(0,1)$. Here $\theta = (\beta, u, \sigma_\varepsilon, \sigma_u, \phi)$ collects all unknown quantities. On the clean data, the posterior of $\beta_{\mathrm{FFREQ}}$ has mean $0.42$ and $95\%$ credible interval $[0.18,\,0.66]$, entirely above zero (Figure~\ref{fig:spatial_posteriors}), suggesting that frequently flooded sites are associated with higher zinc concentrations. A
local regulator could rely on such findings to justify stricter land-use regulations or prohibitions on residential development in flood-prone areas. This creates a clear economic incentive for adversarial intervention.

We posit a stylized attacker (representing, e.g., prospective development interests) whose goal is to neutralize this link. The attacker does not need to specify a full adversarial density: they apply our moment-matching kernel to $h(\theta)=\beta_{\mathrm{FFREQ}}$ with target mean $0$, and may only delete data, so that $w_i \in \{0,1\}$ and $\sum_{i=1}^N (1-w_i) \leq B$.
The weighted posterior is analytically intractable, so this experiment tests whether the attack can operate entirely from posterior samples. The MMD gradient is estimated with Algorithm~\ref{alg:grad_est} from No-U-Turn Sampler (NUTS) draws of the weighted posterior, as implemented in NumPyro \citep{phan2019composable}. To keep the attack computationally feasible, the chains are warm-started across iterations (i.e., see Section~\ref{sec:solution_methods}). The reported posteriors are sampled with separate NUTS runs, independent of those used to construct the attacks.

\begin{figure}[tb]
    \centering \includegraphics[width=0.6\linewidth]{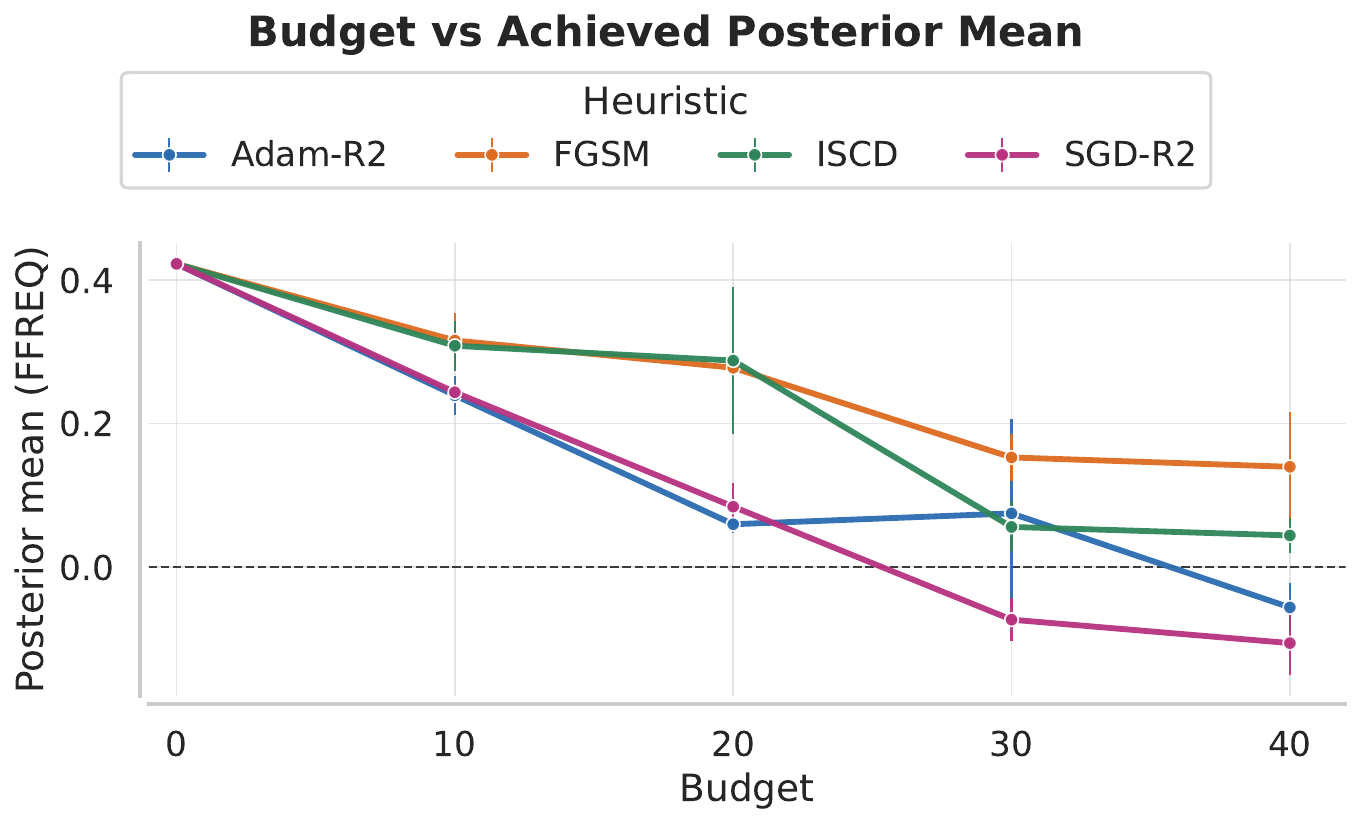}
    \caption{Achieved posterior mean of $\beta_{\mathrm{FFREQ}}$ as a function of the deletion budget $B$. Points are averages over five runs; error bars show $\pm 2$ standard errors.}
    \label{fig:spatial_budget_mean}
\end{figure}

Figure~\ref{fig:spatial_budget_mean} reports the achieved posterior mean of $\beta_{\mathrm{FFREQ}}$ for FGSM, ISCD,\footnote{Here, ISCD forbids undoing earlier moves (a tabu list); see the code repository for details.} Adam-R2, and SGD-R2 over budgets $B \in \{10,20,30,40\}$, with five independent runs per heuristic. With only sampling access to the posterior, the relaxation-based heuristics are the most effective: at budget $20$, Adam-R2 and SGD-R2 are already close to the target, whereas FGSM and ISCD remain much farther away. ISCD, which was competitive on the closed-form objective of Section~\ref{sec:linreg}, now needs $B=30$ to reach the target, whereas FGSM does not reach it using any budget. At larger budgets, the two relaxation-based heuristics can overshoot past $0$ because of the final rounding step required to satisfy the integrality constraints.

We now focus on one of the most effective budget-$20$ runs, obtained by Adam-R2. Figure~\ref{fig:spatial_posteriors} shows that this small budget ($13\%$ of the observations) is already sufficient to nearly neutralize the evidence for a positive flooding effect.
The attack shifts the posterior of $\beta_{\mathrm{FFREQ}}$ toward $0$ and makes sign uncertainty substantial: the posterior mean falls to $0.05$ and the $95\%$ credible interval expands to $[-0.21,\,0.30]$. There is some collateral movement in $\mathrm{DIST\_SQ}$ and $\mathrm{ELEV}$, but the change is clearly most pronounced for $\mathrm{FFREQ}$. Deleting a small fraction of the sites can thus make the posterior treat $\beta_{\mathrm{FFREQ}}$ as effectively irrelevant, potentially softening environmental restrictions on flood-prone land.

Appendix~\ref{app:support_score} examines which sites the attack deletes. Figure~\ref{fig:spatial_attack_map} maps them; they are spread along the river and include both flooding classes, rather than only the most obvious outliers (i.e., frequently flooded sites with very high zinc concentrations, or less frequently flooded sites with very low ones). The attack therefore exploits the interplay between the flooding indicator, the other covariates, and the spatial field, rather than univariate extremes. The appendix also compares the attack with a naive baseline that scores each site by how strongly it supports a positive flooding effect, through its residual under the counterfactual value of the flooding indicator, and deletes the $20$ highest-scoring sites. The optimized attack performs markedly better, moving the posterior of $\beta_{\mathrm{FFREQ}}$ much closer to zero (Figure~\ref{fig:spatial_support_diagnostics}).

\subsection{Attacking Decisions}
\label{sec:radon_decisions}

Our final case study targets a downstream Bayesian decision rather than a set of posterior moments. We use the Minnesota radon survey discussed by \citet{gelman2006data}, which, after dropping incomplete cases, contains radon measurements from $919$ homes in the $85$ counties of Minnesota. For each home, the survey records the county, whether the measurement was taken in the basement or on the first floor, and the radon level in picocuries per liter (pCi/L); for each county, it also records the soil uranium concentration.

We index counties by $j=1,\ldots,J$ and the homes in county $j$ by $i=1,\ldots,n_j$. For home $i$ in county $j$, $y_{ij}$ is the log-radon level and $x_{ij}\in\{0,1\}$ indicates where it was measured ($0$ for the basement, $1$ for the first floor); $u_j$ is the standardized log-uranium level of county $j$. Following \citet{gelman2006data}, we use the hierarchical model
$$
\alpha_j = \mu_\alpha + \beta_{\mathrm{uranium}} u_j + \sigma_\alpha z_j,
\qquad
z_j \sim \mathcal{N}(0,1),
$$
$$
y_{ij} \mid \theta \sim \mathcal{N}\!\bigl(\alpha_j + \beta_{\mathrm{floor}} x_{ij},\; \sigma_y^2\bigr),
$$
with priors $\mu_\alpha \sim \mathcal{N}(0,5^2)$, $\beta_{\mathrm{floor}}, \beta_{\mathrm{uranium}} \sim \mathcal{N}(0,1)$, and $\sigma_\alpha, \sigma_y \sim \mathrm{HalfNormal}(1)$. Here
$
\theta = (\mu_\alpha,\beta_{\mathrm{floor}},\beta_{\mathrm{uranium}},\sigma_\alpha,\sigma_y,z_1,\ldots,z_J)
$. The county intercepts are linked through both the shared uranium trend and the hierarchical shrinkage term $\sigma_\alpha z_j$.

Consider a downstream decision task for a new home in county $j_\star$, whose future log-radon level $\tilde y_\star$ on floor $x_\star$ is unknown. A regulator must make a choice before observing $\tilde y_\star$; they either mandate active remediation measures to lower the radon levels ($d_1$), or take no action ($d_2$). We use the realized utility functions $u(d_1,\tilde y_\star) = -C_R$ and $u(d_2,\tilde y_\star) = -\min\{C_H e^{\tilde y_\star},\, C_{\max}\}$, measured in the same monetary units throughout. The constant $C_R=2000$ is the direct cost of remediation, while $C_H=700$ converts radon exposure $e^{\tilde y_\star}$ into an equivalent health cost. The decision therefore trades off a fixed remediation cost against a radon-dependent exposure cost. We cap the health cost at $C_{\max}=50{,}000$, reached at about $71$ pCi/L (above every measurement in the survey), so that the decision gap defined below is bounded and fulfills the moment conditions of Sections~\ref{sec:targeting_decisions} and~\ref{sec:theory}. Under our model,
$$
\tilde y_\star \mid \theta
\sim \mathcal{N}\!\bigl(\alpha_{j_\star} + \beta_{\mathrm{floor}} x_\star,\; \sigma_y^2\bigr),
$$
where $\pi(\tilde y_\star \mid \theta)$ denotes this density, so $e^{\tilde y_\star}$ is log-normal. Integrating out $\tilde y_\star$ given $\theta$ yields the conditional expected utilities
$$
u(d_1,\theta) = -C_R,
\qquad
u(d_2,\theta) = -\mathbb{E}_{\tilde y_\star \sim \pi(\tilde y_\star \mid \theta)}\bigl[\min\{C_H e^{\tilde y_\star},\, C_{\max}\}\bigr].
$$
Here, $c(\theta) = -u(d_2, \theta)$ defines the conditional \emph{expected health cost}, which has a closed form (Appendix~\ref{app:radon}).
The Bayes action is therefore determined by the sign of the posterior mean of the decision gap
$$
g_1(\theta)
= u(d_1,\theta)-u(d_2,\theta)
= -C_R + c(\theta).
$$
If $\mathbb{E}_{\theta \sim \pi(\theta \mid \mathbf{X})}[g_1(\theta)]>0$, the posterior expected health cost exceeds the remediation cost $C_R$ and remediation is optimal; if it is negative, doing nothing is optimal.

Now consider a strategic attacker, such as a developer, who wants to avoid remediation for a target home that the regulator would remediate under the clean data. The attacker can remove homes from the data used for inference, up to a budget of $B$ homes. To make the attack more subtle and harder to detect, we also assume that the attacker can only remove homes outside the county of the target household. In a hierarchical model, this restriction still leaves room for an attack: partial pooling propagates deletions in other counties, through the shared parameters, to the predictive distribution of the target home. This setup is an instance of the decision-targeting framework of Section~\ref{sec:targeting_decisions}, with target action $d_2$: the attacker seeks $w$ such that $\mathbb{E}_{\theta \sim \pi_w}[g_1(\theta)] < 0$. To enforce this with a safety margin $\gamma = 50$, we take the feature map $\phi(\theta) = g_1(\theta)$ and the target moment $m = -\gamma$.

\begin{figure}[H]
    \centering
    \includegraphics[width=0.56\linewidth]{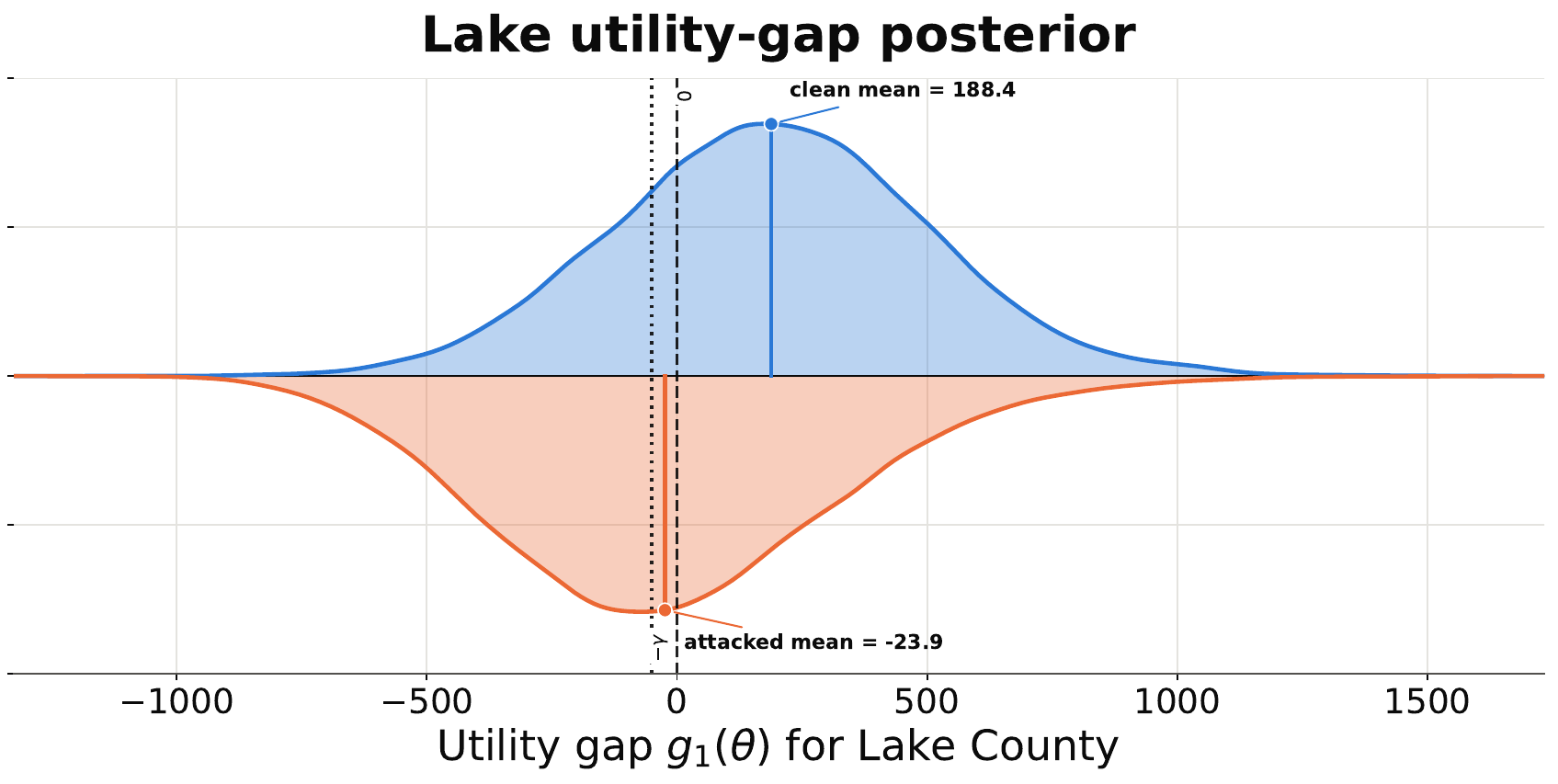}
    \caption{Posterior distribution of the Lake County utility gap $g_1(\theta)$ under the clean data and after the budget-$7$ attack. Positive values favor remediation ($d_1$); negative values favor no action ($d_2)$.}
    \label{fig:radon_gap_posterior}
\end{figure}

We focus on a new home in Lake County and on its basement radon level, i.e., $j_\star=\mathrm{Lake}$ and $x_\star=0$. We choose Lake County because, under the clean posterior, remediation ($d_1$) is optimal, but not by a wide margin, so a small budget may suffice to flip the decision. Indeed, the posterior expected health cost for the target home is about $2188$, above the remediation cost of $2000$. We compute the attack with SGD-R2 (Appendix~\ref{app:radon} gives the settings). Deleting only $7$ homes ($0.76\%$ of the data), all outside Lake County, is enough to flip the Bayes action from $d_1$ (remediate) to $d_2$ (no action). Figure~\ref{fig:radon_gap_posterior} shows the posterior of the Lake County utility gap under the clean and attacked data: its mean moves from about $188$ to $-24$, as the posterior expected health cost drops from about $2188$ to $1976$. The target value $-50$, which includes the safety margin, is not attained, but the Bayes decision is reversed.

The seven deleted homes lie in five other counties. Appendix~\ref{app:radon} shows how their removal propagates through the partially pooled model: Figure~\ref{fig:radon_county_cost} compares the expected health cost of every county before and after the attack, and Figure~\ref{fig:radon_cost_decomposition} splits the change for Lake County into the effects of a lower predicted intercept and a smaller predictive variance.

In summary, this example shows that posterior attraction can be used to manipulate end-to-end Bayesian decisions, not merely posterior summaries. Even when the attacker cannot delete any home in the target county, deleting only seven of the $919$ homes is enough to reverse the optimal action. The same structure that makes hierarchical models statistically efficient under clean data also creates a pathway through which small, well-chosen perturbations can alter downstream decisions.

\section{Conclusion}\label{sec-conc}

Poisoning attacks have been studied extensively for models of the algorithmic culture, but much less in the data-modeling culture, and in Bayesian analysis in particular. This paper helps to close that gap. Building on posterior attraction problems, we have shown how an attacker can alter Bayesian inferences and decisions by deleting or replicating a few observations, chosen to minimize the MMD between the resulting posterior and a target distribution. With a suitable kernel, this target needs to specify only the posterior moments that the attacker cares about, and, since Bayes actions depend on the posterior only through expected utilities, the same approach can target a decision directly.

The attacks are found by gradient-based heuristics, motivated by a convergence analysis of projected stochastic gradient descent on a continuous relaxation of the problem. To decide which observations to delete or replicate, these heuristics need only draws from the weighted posterior. They therefore apply to any Bayesian model whose posterior can be sampled, exactly or by MCMC. In our experiments, small budgets sufficed: deleting $20$ of the $152$ Meuse sites nearly neutralized the evidence for a positive flooding effect, and deleting seven of the $919$ homes in the radon survey reversed the decision to remediate a new home.

These results open several avenues for future research. Posterior attraction problems can be extended along both axes of Section~\ref{sec:pa}. Along the discrepancy axis, using an MMD-based approach instead of the original KL-based one \citep{pmlr-v258-carreau25a} trades the convexity of the latter for the flexibility of the former, as it lets the attacker target a few moments, or a decision, without a full adversarial posterior. Both approaches, however, share a computational bottleneck: every iteration of projected stochastic gradient descent requires fresh posterior draws, typically from MCMC, which is costly for large models and datasets. Other discrepancies that retain these advantages while leading to cheaper or better-behaved optimization problems deserve study. Along the manipulation axis, the constraint set could be made more flexible, for instance by combining deletions and replications with perturbations of the observations \citep{caballero2025posterior}.

Although we have put ourselves in the attacker's shoes, our aim is defensive: vulnerabilities of Bayesian inference must be understood before defenses can be designed. Bayesian defensive frameworks, such as those of \citet{gallego2024protecting} and \citet{arce2025unifying}, model the attacker through an adversarial channel and use this model to build adversarially robust predictive models. These frameworks, however, address evasion attacks, i.e., data manipulation at prediction time. Extending these defenses to data poisoning, using attacks akin to our own as an opponent model, is a natural next step. Finally, beyond the adversarial machine learning perspective adopted herein, the proposed MMD-based attacks may also be used to evaluate the robustness of Bayesian inferences and decisions to the worst-case deletion or replication of data \citep{broderick2023automatic}. Exploring this use, particularly in high-stakes applications, is an interesting direction for future research.

\section{Disclosure statement}\label{disclosure-statement}

The authors report there are no competing interests to declare.

\section{Declaration of generative AI use}\label{ai-declaration}

The authors used Claude Opus 5.5 to polish the writing, as a coding assistant, and to double-check the correctness of the proofs. They reviewed and verified all of its outputs, checked its terms of use, and take full responsibility for the originality, accuracy and integrity of the whole content, including the references.

\section{Data Availability Statement}\label{data-availability-statement}

The data and code that reproduce all experiments are available at \coderepo.

\bibliography{ref.bib}

\if1\fullversion
\newpage
\appendix

\section{Additional Proofs}

This appendix provides a collection of proofs referenced in the main text.

\subsection{Gradient of an expectation under the weighted posterior}
\label{app:proof_grad_cov}
\begin{proof}[Proof of Lemma~\ref{lemma:grad_cov}]
The gradient of the log-density $\log \pi_w(\theta \mid \mathbf{X}) = w^\top f_{\mathbf{X}}(\theta) + \log \pi(\theta) - \log Z(w)$ is given by
\begin{equation} \label{eq:log_density_gradient}
    \nabla_w \log \pi_w(\theta \mid \mathbf{X}) = f_{\mathbf{X}}(\theta) - \nabla_w \log Z(w).
\end{equation}
As a consequence,
\begin{align*}
    \nabla_w \mathbb{E}_{\theta \sim \pi_w}[h(\theta)] &= \int h(\theta) \nabla_w \pi_w(\theta \mid \mathbf{X}) \, d\theta \\
    &= \int h(\theta) \pi_w(\theta \mid \mathbf{X}) \nabla_w \log \pi_w(\theta \mid \mathbf{X}) \, d\theta \\
    &= \mathbb{E}_{\theta \sim \pi_w}[h(\theta) (f_{\mathbf{X}}(\theta) - \nabla_w \log Z(w))].
\end{align*}
Since $\nabla_w \log Z(w) = \mathbb{E}_{\theta' \sim \pi_w}[f_{\mathbf{X}}(\theta')]$, we substitute this back into the expression:
\begin{align*}
    \nabla_w \mathbb{E}_{\theta \sim \pi_w}[h(\theta)] &= \mathbb{E}_{\theta \sim \pi_w}[h(\theta) f_{\mathbf{X}}(\theta)] - \mathbb{E}_{\theta \sim \pi_w}[h(\theta)] \cdot \mathbb{E}_{\theta' \sim \pi_w}[f_{\mathbf{X}}(\theta')].
\end{align*}
By the Cauchy-Schwarz inequality, the assumption that $h(\theta)$ and $\|f_{\mathbf{X}}(\theta)\|_2$ have finite second moments ensures that the expectation of their product $\mathbb{E}_{\theta \sim \pi_w}[h(\theta) f_{\mathbf{X}}(\theta)]$ is finite, and that their individual first moments exist. Therefore, the resulting expression is well-defined and evaluates precisely to the definition of the covariance, completing the proof.
\end{proof}

\subsection{Exact gradient of the squared MMD}
\label{app:proof_obj_grad}
\begin{proof}[Proof of Theorem~\ref{thm:obj_grad}]
Expanding the squared MMD definition and differentiating yields
\begin{equation*}
    \nabla_w \mathbb{E}_{\theta, \theta' \sim \pi_A}[k(\theta, \theta')] - 2\nabla_w \mathbb{E}_{\theta \sim \pi_w, \theta' \sim \pi_A}[k(\theta, \theta')] + \nabla_w \mathbb{E}_{\theta, \theta' \sim \pi_w}[k(\theta, \theta')]
\end{equation*}
\noindent and allows a term-by-term treatment. Since the first term depends solely on the fixed target $\pi_A$, its gradient is zero and can be set aside.

Focusing on the second term, it can be observed that Lemma~\ref{lemma:grad_cov} applies. Namely, let $\mu_A := \mathbb{E}_{\theta' \sim \pi_A}[k(\theta', \cdot)]$ be the kernel mean embedding of $\pi_A$ in the RKHS $\mathcal{H}_k$, and note that
\begin{equation*}
    -2\nabla_w \mathbb{E}_{\theta \sim \pi_w, \theta' \sim \pi_A}[k(\theta, \theta')] = -2\nabla_w \mathbb{E}_{\theta \sim \pi_w} \left[ h_{\pi_A}(\theta) \right],
\end{equation*}
where $h_{\pi_A}(\theta) := \mathbb{E}_{\theta' \sim \pi_A}[k(\theta, \theta')]$ is the mean embedding evaluated at $\theta$. By the reproducing property and the linearity of the inner product, $h_{\pi_A}(\theta) = \langle k(\theta, \cdot), \mu_A \rangle_{\mathcal{H}_k}$. Therefore,
\begin{equation*}
h_{\pi_A}(\theta)^2 \le \Vert k(\theta, \cdot)\|_{\mathcal{H}_k}^2 \|\mu_A\|_{\mathcal{H}_k}^2 = k(\theta, \theta) \|\mu_A\|_{\mathcal{H}_k}^2
\end{equation*}
\noindent where the inequality results from the Cauchy-Schwarz inequality in $\mathcal{H}_k$, and the equality derives from the reproducing property. The coupled use of Jensen's inequality with the reproducing property yields $\|\mu_A\|_{\mathcal{H}_k}^2 \le\mathbb{E}_{\theta' \sim \pi_A}[k(\theta', \theta')] < \infty$ which is finite by assumption. As such,
\begin{equation*}
\mathbb{E}_{\theta \sim \pi_w}[h_{\pi_A}(\theta)^2] \le \|\mu_A\|_{\mathcal{H}_k}^2 \mathbb{E}_{\theta \sim \pi_w}[k(\theta, \theta)]
\end{equation*}
\noindent which is also finite by assumption. This result, coupled with the assumption on $\|f_{\mathbf{X}}(\theta)\|_2$, implies the gradient of the second term is $-2\mathrm{Cov}_{\pi_w}(h_{\pi_A}(\theta), f_{\mathbf{X}}(\theta))$.

Similar results hold for the third term. Namely, this gradient can be re-expressed as
\begin{align*}
    \nabla_w \mathbb{E}_{\theta, \theta' \sim \pi_w}[k(\theta, \theta')] &= \iint k(\theta, \theta') \left( [\nabla_w \pi_w(\theta\mid  \mathbf{X})] \pi_w(\theta'\mid  \mathbf{X}) + \pi_w(\theta\mid  \mathbf{X} ) [\nabla_w \pi_w(\theta'\mid  \mathbf{X})] \right) d\theta d\theta'\\
    &=2 \int \left[ \int k(\theta, \theta') \pi_w(\theta'\mid  \mathbf{X}) d \theta' \right] \pi_w(\theta\mid  \mathbf{X}) \nabla_w \log \pi_w(\theta \mid  \mathbf{X}) d \theta \\
    &= 2 \mathbb{E}_{\theta \sim \pi_w}[h_{\pi_w}(\theta) \nabla_w \log \pi_w(\theta\mid  \mathbf{X})].
\end{align*}
\noindent where the first equality is justified by the assumption on differentiation--integration interchange, the second by the REINFORCE trick combined with symmetry of the kernel and distributions, and the third by defining $h_{\pi_w}(\theta) := \mathbb{E}_{\theta' \sim \pi_w}[k(\theta, \theta')]$. By substituting the log-density gradient from equation \eqref{eq:log_density_gradient} and recalling that $\nabla_w \log Z(w) = \mathbb{E}_{\theta' \sim \pi_w}[f_{\mathbf{X}}(\theta')]$, it follows that
\begin{equation*}
    \nabla_w \mathbb{E}_{\theta, \theta' \sim \pi_w}[k(\theta, \theta')] = 2 \cdot \mathrm{Cov}_{\pi_w}(h_{\pi_w}(\theta), f_{\mathbf{X}}(\theta)).
\end{equation*}
As in the analysis of $h_{\pi_A}(\theta)$, the assumption $\mathbb{E}_{\theta \sim \pi_w}[k(\theta, \theta)] < \infty$ ensures that $h_{\pi_w}(\theta)$ also has a finite second moment under $\pi_w$. Together with the assumption that $\|f_{\mathbf{X}}(\theta)\|_2$ has a finite second moment, this guarantees that all integrability requirements are met and the covariance is well-defined.

Summing the non-zero gradients, and using the linearity of the covariance, we obtain
\begin{align*}
    \nabla_w \mathrm{MMD}^2(\pi_A, \pi_w; k) &= 2 \cdot \mathrm{Cov}_{\pi_w}(h_{\pi_w}(\theta), f_{\mathbf{X}}(\theta)) - 2 \cdot \mathrm{Cov}_{\pi_w}(h_{\pi_A}(\theta), f_{\mathbf{X}}(\theta)) \\
    &= 2 \cdot \mathrm{Cov}_{\pi_w}\left( h_{\pi_w}(\theta) - h_{\pi_A}(\theta), f_{\mathbf{X}}(\theta) \right).
\end{align*}
Substituting $g(\theta) = h_{\pi_w}(\theta) - h_{\pi_A}(\theta)$ completes the proof.
\end{proof}

\subsection{Curvature of the MMD-squared objective}
\label{app:curvature}

\noindent\textit{Proof of Lemma \ref{lemma:Lsmooth}.}
We compute the first two derivatives of $F$, bound its Hessian on $\mathcal{W}$, and deduce that $\nabla F$ is Lipschitz. Write $F(w) = \mathrm{MMD}^2(\pi_A, \pi_w; k) = \|\mu_w - \mu_A\|_{\mathcal{H}_k}^2$ as $F = \Psi \circ \mu$, with $\mu(w) = \mu_w$ and $\Psi(h) = \|h - \mu_A\|_{\mathcal{H}_k}^2$, which is well defined because $\mu_A \in \mathcal{H}_k$. The map $\Psi$ is smooth on $\mathcal{H}_k$, with $D\Psi(h)[v] = 2\langle h - \mu_A, v \rangle_{\mathcal{H}_k}$ and $D^2\Psi(h)[v, v'] = 2\langle v, v' \rangle_{\mathcal{H}_k}$, and $\mu$ is twice continuously Fr\'echet differentiable on an open neighborhood $U$ of $\mathcal{W}$ by assumption; hence $F$ is twice continuously differentiable on $U$. Writing $\partial_i \mu_w = D\mu(w)[e_i]$ and $\partial^2_{ij} \mu_w = D^2 \mu(w)[e_i, e_j]$ for the partial derivatives of $\mu_w$, where $e_i$ is the $i$th canonical basis vector, $\partial_i = \partial/\partial w_i$ and $\partial^2_{ij} = \partial^2/\partial w_i \partial w_j$, the chain rule gives
\begin{align*}
\partial_i F(w) &= D\Psi(\mu_w)[\partial_i \mu_w] = -2 \langle \mu_A - \mu_w, \partial_i \mu_w \rangle_{\mathcal{H}_k}, \\
\partial^2_{ij} F(w) &= D^2\Psi(\mu_w)[\partial_i \mu_w, \partial_j \mu_w] + D\Psi(\mu_w)[\partial^2_{ij} \mu_w] \\
&= 2 \langle \partial_i \mu_w, \partial_j \mu_w \rangle_{\mathcal{H}_k} - 2 \langle \mu_A - \mu_w, \partial^2_{ij} \mu_w \rangle_{\mathcal{H}_k}.
\end{align*}
In matrix form, with the inner products taken component-wise,
\begin{align}
\nabla_w F(w) &= -2 \langle \mu_A - \mu_w, \nabla_w \mu_w \rangle_{\mathcal{H}_k}, \label{eq:mmd_grad_embed} \\
\nabla^2_w F(w) &= 2 \langle \nabla_w \mu_w, \nabla_w \mu_w \rangle_{\mathcal{H}_k} - 2 \langle \mu_A - \mu_w, \nabla^2_w \mu_w \rangle_{\mathcal{H}_k}. \label{eq:mmd_hess_embed}
\end{align}

Next, we bound the Hessian on $\mathcal{W}$. Since $F$ is twice continuously differentiable on $U$, its Hessian $w \mapsto \nabla^2_w F(w)$ is continuous in operator norm. As $\mathcal{W}$ is closed and bounded, hence compact, the extreme value theorem gives $\xi = \sup_{w \in \mathcal{W}} \|\nabla^2_w F(w)\|_{\mathrm{op}} < \infty$. Because the Hessian is symmetric, this is the two-sided bound $-\xi I \preceq \nabla^2_w F(w) \preceq \xi I$, which controls the magnitude of the curvature but not its sign.

Finally, because $\mathcal{W}$ is convex, the segment between any $v, w \in \mathcal{W}$ stays in $\mathcal{W}$, where $\|\nabla^2_w F\|_{\mathrm{op}} \le \xi$. The mean value inequality for $\nabla F$ along this segment then gives $\|\nabla F(v) - \nabla F(w)\|_2 \le \xi \|v - w\|_2$, so $F$ is $\xi$-smooth on $\mathcal{W}$.\hfill$\square$

\medskip

\noindent \textit{Remark.} When differentiation and integration can be interchanged, the score identity $\partial \pi_w / \partial w_i = s_i \, \pi_w$, with $s_i(\theta) = \log \pi(X_i \mid \theta) - \mathbb{E}_{\theta^* \sim \pi_w}[\log \pi(X_i \mid \theta^*)]$, gives $\partial_i \mu_w = \mathbb{E}_{\theta \sim \pi_w}[s_i(\theta)\, k(\theta, \cdot)]$; by the reproducing property, \eqref{eq:mmd_grad_embed} then reduces to the covariance formula of Theorem~\ref{thm:obj_grad}.

\subsection{The gradient estimator is unbiased}
\label{app:unbiased}

\noindent\textit{Proof of Lemma \ref{lemma:grad_unbiased}.} Fix $w \in \mathcal{W}$. Recall that
\begin{equation*}
\hat{g}(\theta) = \frac{1}{N} \sum_{\theta' \in S_{\text{est}}} k(\theta, \theta') - \frac{1}{M} \sum_{\theta' \in S_A} k(\theta, \theta'),
\end{equation*}
and let
\begin{equation*}
\widehat{\mathrm{Cov}} = \frac{1}{N-1} \sum_{\theta \in S_{\text{cov}}} \left( \hat{g}(\theta) - \bar{g} \right) \left( f_{\mathbf{X}}(\theta) - \bar{f} \right),
\end{equation*}
where $\bar{g} = \frac{1}{N} \sum_{\theta \in S_{\text{cov}}} \hat{g}(\theta)$ and $\bar{f} = \frac{1}{N} \sum_{\theta \in S_{\text{cov}}} f_{\mathbf{X}}(\theta)$. By construction $S_{\text{est}} \sim \pi_w^N$, $S_A \sim \pi_A^M$, and $S_{\text{cov}} \sim \pi_w^N$, so that
\begin{align*}
\mathbb{E}_{S_{\text{est}}, S_A, S_{\text{cov}}}[\hat{\nabla}]
&= 2 \, \mathbb{E}_{S_{\text{est}}, S_A}\!\left[ \mathbb{E}_{S_{\text{cov}}}\!\left[ \widehat{\mathrm{Cov}} \,\middle|\, S_{\text{est}}, S_A \right] \right]
= 2 \, \mathbb{E}_{S_{\text{est}}, S_A}\!\left[ \mathrm{Cov}_{\pi_w}(\hat{g}(\theta), f_{\mathbf{X}}(\theta)) \right].
\end{align*}
Substituting the definition of the covariance and exchanging the order of expectation gives
\begin{align*}
\mathbb{E}_{S_{\text{est}}, S_A, S_{\text{cov}}}[\hat{\nabla}]
&= 2 \, \mathbb{E}_{\theta \sim \pi_w}\!\left[ \mathbb{E}_{S_{\text{est}}, S_A}\!\left[ \hat{g}(\theta) - \mathbb{E}_{\theta \sim \pi_w}[\hat{g}(\theta)] \right] \left( f_{\mathbf{X}}(\theta) - \mathbb{E}_{\theta \sim \pi_w}[f_{\mathbf{X}}(\theta)] \right) \right] \\
&= 2 \, \mathbb{E}_{\theta \sim \pi_w}\!\left[ \left( g(\theta) - \mathbb{E}_{\theta \sim \pi_w}[g(\theta)] \right) \left( f_{\mathbf{X}}(\theta) - \mathbb{E}_{\theta \sim \pi_w}[f_{\mathbf{X}}(\theta)] \right) \right] \\
&= 2 \, \mathrm{Cov}_{\pi_w}\!\left( g(\theta), f_{\mathbf{X}}(\theta) \right) = \nabla_w \mathrm{MMD}^2(\pi_A, \pi_w; k),
\end{align*}
where the last line invokes Theorem \ref{thm:obj_grad}. At iteration $t$ of projected SGD, $w_t$ is $\mathcal{F}_t$-measurable and, conditionally on $\mathcal{F}_t$, the three batches are fresh i.i.d.\ draws from $\pi_{w_t}$ and $\pi_A$; applying this identity with $w = w_t$ therefore gives $\mathbb{E}[\hat{\nabla}_t \mid \mathcal{F}_t] = \nabla F(w_t)$. \hfill $\square$

\subsection{Bounded conditional variance of the stochastic gradient}
\label{app:variance}

\noindent\textit{Proof of Lemma \ref{lemma:gradient_normsq}.} Fix an iterate $w = w_t$ and write $\mathbb{E}[\,\cdot\,]$ for the conditional expectation $\mathbb{E}[\,\cdot \mid \mathcal{F}_t]$, that is, the expectation over the mutually independent fresh batches $S_{\text{est}}, S_{\text{cov}} \sim \pi_w^N$ and $S_A \sim \pi_A^M$ drawn at that iteration (Appendix~\ref{app:unbiased}). Write $\theta_1, \dots, \theta_N$ for the elements of $S_{\text{cov}}$ and set $Y_a = \hat{g}(\theta_a) - \bar{g}$ an
 d $Z_{ia} = (f_{\mathbf{X}}(\theta_a) - \bar{f})_i$, so that the $i$th coordinate of $\hat{\nabla}$ is $\frac{2}{N-1} \sum_{a=1}^N Y_a Z_{ia}$.

We first bound the fourth moments of $Y_a$ and $Z_{ia}$ uniformly over $\mathcal{W}$. Let
\begin{gather*}
\kappa_{\mathcal{W}} := \sup_{w \in \mathcal{W}} \mathbb{E}_{\theta \sim \pi_w}[k(\theta,\theta)^2], \qquad \kappa_A := \mathbb{E}_{\theta \sim \pi_A}[k(\theta,\theta)^2], \\
V := \frac{1}{N}\sum_{\theta' \in S_{\text{est}}} \sqrt{k(\theta',\theta')} + \frac{1}{M}\sum_{\theta' \in S_A} \sqrt{k(\theta',\theta')},
\end{gather*}
where $\kappa_{\mathcal{W}}$ and $\kappa_A$ are finite by hypothesis. By the Cauchy--Schwarz inequality in $\mathcal{H}_k$, $|k(\theta,\theta')| \le \sqrt{k(\theta,\theta)\,k(\theta',\theta')}$, so $|\hat{g}(\theta)| \le \sqrt{k(\theta,\theta)}\, V$. The inequality $(x+y)^4 \le 8(x^4+y^4)$ and Jensen's inequality give $\mathbb{E}[V^4] \le 8(\kappa_{\mathcal{W}} + \kappa_A)$, and since $V$ is independent of $S_{\text{cov}}$, $\mathbb{E}[\hat{g}(\theta_a)^4] \le \kappa_{\mathcal{W}}\, \mathbb{E}[V^4] \le 8\kappa_{\mathcal{W}}(\kappa_{\mathcal{W}} + \kappa_A)$ for every $a$. To account for the centering at the sample means $\bar{g}$ and $\bar{f}$, note that, for any random variables $U_1, \dots, U_N$ with $\bar{U} = N^{-1} \sum_{b} U_b$, the same two inequalities give, without any independence,
\begin{align*}
\mathbb{E}[(U_a - \bar{U})^4]
&\le 8\bigl(\mathbb{E}[U_a^4] + \mathbb{E}[\bar{U}^4]\bigr)
\le 8\Bigl(\mathbb{E}[U_a^4] + \frac{1}{N} \sum_{b=1}^N \mathbb{E}[U_b^4]\Bigr)
\le 16 \max_{b} \mathbb{E}[U_b^4].
\end{align*}
Applying this to $U_b = \hat{g}(\theta_b)$ and to $U_b = \log \pi(X_i \mid \theta_b)$ yields
\begin{align*}
\mathbb{E}[Y_a^4] &\le 128\,\kappa_{\mathcal{W}}(\kappa_{\mathcal{W}} + \kappa_A) =: C_Y^2, \\
\mathbb{E}[Z_{ia}^4] &\le 16 \sup_{w \in \mathcal{W}} \max_{1 \le j \le n} \mathbb{E}_{\theta \sim \pi_w}\!\left[\bigl(\log \pi(X_j \mid \theta)\bigr)^4\right] =: C_Z^2,
\end{align*}
where $C_Z < \infty$ by the uniform moment bound on the log-likelihoods (Section~\ref{sec:theory}).

Next, the inequality $\bigl(\sum_{a=1}^N x_a\bigr)^2 \le N \sum_{a=1}^N x_a^2$ and the Cauchy--Schwarz inequality for expectations, neither of which requires independence, give
\begin{align*}
\mathbb{E}\bigl[\|\hat{\nabla}\|_2^2\bigr]
&= \frac{4}{(N-1)^2} \sum_{i=1}^n \mathbb{E}\Bigl[\Bigl(\sum_{a=1}^N Y_a Z_{ia}\Bigr)^2\Bigr]
\le \frac{4N}{(N-1)^2} \sum_{i=1}^n \sum_{a=1}^N \mathbb{E}\bigl[Y_a^2 Z_{ia}^2\bigr] \\
&\le \frac{4N}{(N-1)^2} \sum_{i=1}^n \sum_{a=1}^N \sqrt{\mathbb{E}[Y_a^4]\, \mathbb{E}[Z_{ia}^4]}
\le \frac{4 n N^2 C_Y C_Z}{(N-1)^2} =: \sigma^2 < \infty.
\end{align*}
Since $\mathbb{E}$ is conditional on $\mathcal{F}_t$, this reads $\mathbb{E}[\|\hat{\nabla}_t\|_2^2 \mid \mathcal{F}_t] \le \sigma^2$. As $\hat{\nabla}_t$ is conditionally unbiased (Lemma~\ref{lemma:grad_unbiased}), it follows that
\begin{equation*}
\mathbb{E}\bigl[\|\hat{\nabla}_t - \nabla F(w_t)\|_2^2 \mid \mathcal{F}_t\bigr] = \mathbb{E}\bigl[\|\hat{\nabla}_t\|_2^2 \mid \mathcal{F}_t\bigr] - \|\nabla F(w_t)\|_2^2 \le \sigma^2.
\end{equation*}
Neither $C_Y$ nor $C_Z$ depends on $w_t$, so both bounds hold uniformly over $\mathcal{W}$.\hfill$\square$

\subsection{Finite-time stationarity bound for projected SGD}
\label{app:psgd}
We prove Theorem \ref{thm:psgd_stationary} by specializing Theorem 2(a) and Corollary 3 of \citet{ghadimi2016mini} to the Euclidean projected setting, with feasible set $\mathcal{W}$, no nonsmooth term, smooth part $F$, and Lipschitz constant $\xi$. Recall from Theorem~\ref{thm:psgd_stationary} that the step size is fixed at $\eta = 1/(2\xi)$, $F^\star := \inf_{w \in \mathcal{W}} F(w)$ denotes the infimum of the objective over the feasible set, and $\nu_t^2$ is the conditional-variance bound of the estimator, $\mathbb{E}[\|\hat{\nabla}_t - \nabla F(w_t)\|_2^2 \mid \mathcal{F}_t] \le \nu_t^2$, as assumed in the theorem; Lemma~\ref{lemma:gradient_normsq} permits the choice $\nu_t^2 = \sigma^2$. It is convenient to introduce the \emph{stochastic} gradient mapping $\widetilde{G}_t$, the gradient error $\delta_t$, and the true gradient mapping $G_t$,
\begin{equation*}
\widetilde{G}_t := \frac{1}{\eta}(w_t - w_{t+1}), \qquad
\delta_t := \hat{\nabla}_t - \nabla F(w_t), \qquad
G_t := G_\eta(w_t),
\end{equation*}
so that the update reads $w_{t+1} - w_t = -\eta \widetilde{G}_t$.

We begin by noting that the $\xi$-smoothness of $F$ implies
\begin{equation*}
F(w_{t+1}) \le F(w_t) + \nabla F(w_t)^\top (w_{t+1} - w_t) + \frac{\xi}{2} \| w_{t+1} - w_t \|_2^2
= F(w_t) - \eta \nabla F(w_t)^\top \widetilde{G}_t + \frac{\xi \eta^2}{2} \| \widetilde{G}_t \|_2^2.
\end{equation*}
Writing $\nabla F(w_t) = \hat{\nabla}_t - \delta_t$,
\begin{equation*}
F(w_{t+1}) \le F(w_t) - \eta \, \hat{\nabla}_t^\top \widetilde{G}_t + \eta \, \delta_t^\top \widetilde{G}_t + \frac{\xi \eta^2}{2} \| \widetilde{G}_t \|_2^2.
\end{equation*}

To control the inner product $\hat{\nabla}_t^\top \widetilde{G}_t$ we invoke the defining optimality of the projection. Since $w_{t+1} = \Pi_\mathcal{W}(w_t - \eta \hat{\nabla}_t) = \argmin_{u \in \mathcal{W}} \{ \hat{\nabla}_t^\top u + \frac{1}{2\eta}\|u - w_t\|_2^2 \}$, the first-order optimality condition gives, for all $u \in \mathcal{W}$,
\begin{equation*}
\big(\hat{\nabla}_t + \tfrac{1}{\eta}(w_{t+1} - w_t)\big)^\top (u - w_{t+1}) \ge 0.
\end{equation*}
Taking $u = w_t$ and using $\tfrac{1}{\eta}(w_{t+1} - w_t) = -\widetilde{G}_t$ and $w_t - w_{t+1} = \eta \widetilde{G}_t$ yields $\eta \, (\hat{\nabla}_t - \widetilde{G}_t)^\top \widetilde{G}_t \ge 0$, hence $\hat{\nabla}_t^\top \widetilde{G}_t \ge \| \widetilde{G}_t \|_2^2$. Substituting,
\begin{equation*}
F(w_{t+1}) \le F(w_t) - \left( \eta - \frac{\xi \eta^2}{2} \right) \| \widetilde{G}_t \|_2^2 + \eta \, \delta_t^\top \widetilde{G}_t.
\end{equation*}

It remains to handle the noise term $\delta_t^\top \widetilde{G}_t$ in conditional expectation given $\mathcal{F}_t$, under which the iterate $w_t$, and hence the deterministic mapping $G_t$, is fixed. We split $\delta_t^\top \widetilde{G}_t = \delta_t^\top G_t + \delta_t^\top (\widetilde{G}_t - G_t)$. The first piece vanishes in conditional expectation, because $G_t$ is $\mathcal{F}_t$-measurable and $\mathbb{E}[\delta_t \mid \mathcal{F}_t] = 0$ by unbiasedness (Lemma \ref{lemma:grad_unbiased}), so $\mathbb{E}[\delta_t^\top G_t \mid \mathcal{F}_t] = 0$. For the second piece, the two mappings differ only through the gradient vector supplied to the projection, and the Euclidean projection is nonexpansive, so
\begin{equation*}
\| \widetilde{G}_t - G_t \|_2 = \frac{1}{\eta} \left\| \Pi_\mathcal{W}(w_t - \eta \nabla F(w_t)) - \Pi_\mathcal{W}(w_t - \eta \hat{\nabla}_t) \right\|_2 \le \| \hat{\nabla}_t - \nabla F(w_t) \|_2 = \| \delta_t \|_2.
\end{equation*}
By the Cauchy--Schwarz inequality, $\delta_t^\top (\widetilde{G}_t - G_t) \le \| \delta_t \|_2 \, \| \widetilde{G}_t - G_t \|_2 \le \| \delta_t \|_2^2$. Taking conditional expectations given $\mathcal{F}_t$ and using the variance bound $\mathbb{E}[\|\delta_t\|_2^2 \mid \mathcal{F}_t] \le \nu_t^2$,
\begin{align*}
\mathbb{E}[ F(w_{t+1}) \mid \mathcal{F}_t ] &\le F(w_t) - \left( \eta - \frac{\xi \eta^2}{2} \right) \mathbb{E}[ \| \widetilde{G}_t \|_2^2 \mid \mathcal{F}_t ] + \eta \, \nu_t^2 \\
&\le F(w_t) - \left( \eta - \xi \eta^2\right) \mathbb{E}[ \| \widetilde{G}_t \|_2^2 \mid \mathcal{F}_t ] + \eta \, \nu_t^2.
\end{align*}
In the last inequality, we replace $\eta - \xi\eta^2/2$ by the smaller positive coefficient $\eta - \xi\eta^2 = 1/(4\xi)$ to obtain the stated constants.

We now remove the conditioning and accumulate over iterations. Taking total expectations through the tower property, $\mathbb{E}[\,\cdot\,] = \mathbb{E}[\mathbb{E}[\,\cdot \mid \mathcal{F}_t]]$, and summing the resulting inequality over $t = 0,\dots,T-1$, the objective differences telescope into $\mathbb{E}[F(w_T)] - F(w_0)$; bounding the final iterate from below by $F(w_T) \ge F^\star$ and rearranging leaves
\begin{equation*}
(\eta - \xi \eta^2) \sum_{t=0}^{T-1} \mathbb{E}\| \widetilde{G}_t \|_2^2 \le F(w_0) - F^\star + \eta \sum_{t=0}^{T-1} \nu_t^2.
\end{equation*}
Since $\eta = 1/(2\xi)$, we have $\eta - \xi \eta^2 = 1/(4\xi)$ and $\eta / (\eta - \xi\eta^2) = 2$. Dividing through by $(\eta - \xi\eta^2)\,T$ to turn the running sum into an iteration average then gives
\begin{equation*}
\frac{1}{T} \sum_{t=0}^{T-1} \mathbb{E}\| \widetilde{G}_t \|_2^2 \le \frac{4 \xi (F(w_0) - F^\star)}{T} + \frac{2}{T} \sum_{t=0}^{T-1} \nu_t^2.
\end{equation*}

The bound just obtained controls the \emph{stochastic} mapping $\widetilde{G}_t$; we convert it into one on the \emph{true} mapping $G_t$ reported by the theorem. Using $\|a + b\|_2^2 \le 2\|a\|_2^2 + 2\|b\|_2^2$ with $a = \widetilde{G}_t$ and $b = G_t - \widetilde{G}_t$, together with the projection bound $\|G_t - \widetilde{G}_t\|_2 \le \|\delta_t\|_2$ established above and $\mathbb{E}[\|\delta_t\|_2^2 \mid \mathcal{F}_t] \le \nu_t^2$,
\begin{equation*}
\mathbb{E}\| G_t \|_2^2 \le 2\, \mathbb{E}\| \widetilde{G}_t \|_2^2 + 2\, \mathbb{E}\| G_t - \widetilde{G}_t \|_2^2 \le 2\, \mathbb{E}\| \widetilde{G}_t \|_2^2 + 2 \nu_t^2.
\end{equation*}
Averaging this over $t$ and inserting the previous display gives
\begin{equation*}
\frac{1}{T} \sum_{t=0}^{T-1} \mathbb{E}\| G_t \|_2^2 \le \frac{8 \xi (F(w_0) - F^\star)}{T} + \frac{6}{T} \sum_{t=0}^{T-1} \nu_t^2.
\end{equation*}
Finally, because the output index $R$ is drawn uniformly from $\{0,\dots,T-1\}$ independently of the algorithm, the expected squared mapping at $w_R$ equals the iteration average, $\mathbb{E}\| G_\eta(w_R) \|_2^2 = T^{-1} \sum_{t=0}^{T-1} \mathbb{E}\| G_t \|_2^2$. This is exactly the bound \eqref{eq:stationarity_bound}, which completes the proof. \hfill$\square$

\subsection{The convex case}
\label{app:convex}

When the objective happens to be convex, the conclusion of Theorem \ref{thm:psgd_stationary} strengthens considerably, since stationarity is then no longer merely necessary for optimality. If $F$ is convex on $\mathcal{W}$, every first-order stationary point is a global minimizer of the continuous relaxation, so the theorem already delivers (approximate) global optimality. If $F$ is moreover $c$-strongly convex, a stronger and more classical conclusion becomes available. The stochastic-approximation analysis of \citet{nemirovski2009robust} and \citet{shapiro2021lectures} then guarantees mean-square convergence to the continuous optimum $w^c$, provided three conditions hold: the step size follows the decreasing schedule $\eta_t = 1/[c(t+1)]$, $t \ge 0$; the gradient estimator is conditionally unbiased (Lemma \ref{lemma:grad_unbiased}); and its second moment is uniformly bounded by some $\sigma^2 < \infty$ (Lemma \ref{lemma:gradient_normsq}). Under these conditions the iterates obey, for every $t \ge 0$,
\begin{equation*}
\mathbb{E}\left[\|w_t - w^c\|_2^2\right] \le \frac{1}{t+1} \max\left\{ \frac{\sigma^2}{c^2}, \ \|w_0 - w^c\|_2^2 \right\}.
\end{equation*}
In the notation of Lemma \ref{lemma:Lsmooth}, strong convexity amounts to the Hessian admitting a uniform positive lower bound; the precise condition is stated next.

\begin{lemma} \label{lemma:strong-convex}
Under the regularity conditions of Lemma \ref{lemma:Lsmooth}, $F = \mathrm{MMD}^2(\pi_A, \pi_w; k)$ is $c$-strongly convex on $\mathcal{W}$ whenever
\begin{equation*}
\langle \nabla_w \mu_w, \nabla_w \mu_w \rangle_{\mathcal{H}_k} \succeq \langle \mu_A - \mu_w, \nabla^2_w \mu_w \rangle_{\mathcal{H}_k} + \frac{c}{2} I, \qquad \forall w \in \mathcal{W},
\end{equation*}
with the inner products taken component-wise and $I$ the $n$-dimensional identity.
\end{lemma}

\noindent\textit{Proof.}
A twice-differentiable function is $c$-strongly convex on the convex set $\mathcal{W}$ if $\nabla^2_w F(w) \succeq c I$ for all $w \in \mathcal{W}$. By \eqref{eq:mmd_hess_embed} this Hessian bound holds exactly when the condition of the lemma does.\hfill$\square$

Equation \eqref{eq:mmd_hess_embed} for the Hessian makes the asymmetry between the two requirements transparent. The Hessian is the sum of a \emph{data term} $2\langle \nabla_w \mu_w, \nabla_w \mu_w \rangle$, which is positive semidefinite, and a \emph{residual term} $-2\langle \mu_A - \mu_w, \nabla^2_w \mu_w \rangle$, whose sign is indefinite and which need not be dominated by the data term. Smoothness asks only that the \emph{magnitude} of this sum be uniformly bounded, a two-sided condition that constrains neither its sign nor its definiteness and is therefore mild. Strong convexity, by contrast, asks the sum to remain uniformly \emph{above} $cI$, so that the positive data term must dominate the indefinite residual term by a fixed margin at every $w \in \mathcal{W}$. This is a substantially more demanding requirement and, for MMD objectives, may fail even where smoothness holds comfortably; this is why we take stationarity, rather than global optimality, as the default guarantee.

While strong convexity remains a theoretical possibility for infinite-dimensional characteristic kernels like the RBF, it generally fails for the moment-matching and decision kernels of Section~\ref{sec:mpa}. Writing $J_w$ for the Jacobian of $w \mapsto \mathbb{E}_{\theta \sim \pi_w}[\phi(\theta)]$, the data term of these linear kernels equals $2 J_w^\top J_w$, whose rank is at most the number of features, typically much smaller than $n$. When there are fewer features than observations, at any $w \in \mathcal{W}$ that attains the targeted moments the residual term vanishes, so the Hessian is singular there and no $c > 0$ satisfies the sufficient condition of Lemma \ref{lemma:strong-convex}. This is why first-order stationarity (Theorem \ref{thm:psgd_stationary}), rather than global strong convexity, serves as the primary performance guarantee for our targeted posterior attraction framework.

A sharper version of the rounding bound \eqref{eq:rounding_distortion} is available in the convex case, where the distortion can be measured against the \emph{integer} optimum itself. Let $w^c$ denote a continuous minimizer of $F$ over $\mathcal{W}$, $w^*_{int} \in \argmin_{w \in \mathbb{Z}^n_{\ge 0} \cap \mathcal{W}} F(w)$ an integer optimum, and $\hat{w} = \Pi_{\mathbb{Z}^n_{\ge 0} \cap \mathcal{W}}(w^c)$ the rounded continuous minimizer.

\begin{corollary} \label{cor:integrality_gap}
Suppose $F$ is $\xi$-smooth on $\mathcal{W}$ (Lemma \ref{lemma:Lsmooth}) and let $w^c$ be a global minimizer of $F$ over $\mathcal{W}$, the point targeted by projected SGD when $F$ is convex. Writing $r(w^c, \hat{w}) := \nabla F(w^c)^\top (\hat{w} - w^c) + \tfrac{\xi}{2}\|\hat{w} - w^c\|_2^2$, the absolute and relative optimality gaps of the rounded solution satisfy
\begin{equation*}
F(\hat{w}) - F(w^*_{int}) \le r(w^c, \hat{w}), \qquad \frac{F(\hat{w}) - F(w^*_{int})}{F(w^*_{int})} \le \frac{r(w^c, \hat{w})}{F(w^c)},
\end{equation*}
the relative bound provided $F(w^c) \neq 0$.
\end{corollary}

\noindent\textit{Proof.}
Because the integer feasible set is contained in the relaxed one, $\mathbb{Z}^n_{\ge 0} \cap \mathcal{W} \subseteq \mathcal{W}$, the three values are ordered as $F(w^c) \le F(w^*_{int}) \le F(\hat{w})$. Applying the quadratic bound \eqref{eq:rounding_distortion} with $w^c$ in place of $w_T$,
\begin{equation*}
F(w^*_{int}) - F(w^c) \le F(\hat{w}) - F(w^c) \le r(w^c, \hat{w}).
\end{equation*}
At the minimizer $w^c$, first-order optimality gives $\nabla F(w^c)^\top(\hat{w} - w^c) \ge 0$ for the feasible point $\hat{w}$, so $r(w^c, \hat{w}) \ge 0$. Rearranging the chain $F(\hat{w}) \le F(w^c) + r(w^c, \hat{w}) \le F(w^*_{int}) + r(w^c, \hat{w})$ yields the absolute gap $F(\hat{w}) - F(w^*_{int}) \le r(w^c, \hat{w})$; dividing by $F(w^*_{int}) \ge F(w^c)$, and using $r(w^c, \hat{w}) \ge 0$ with $F(w^c) \ne 0$, yields the relative gap $\bigl(F(\hat{w}) - F(w^*_{int})\bigr)/F(w^*_{int}) \le r(w^c, \hat{w})/F(w^c)$. Outside the convex case a global minimizer $w^c$ is not available in practice and only the local bound \eqref{eq:rounding_distortion} applies.\hfill$\square$

\section{Targeting Continuous Decisions}\label{app:continuous}

The decision-targeting construction of Section~\ref{sec:targeting_decisions} extends to continuous decisions. Let $\mathcal{D} \subseteq \mathbb{R}^p$ be a convex action set and suppose $u(d, \theta)$ is continuously differentiable and strictly concave in $d$ for every $\theta \in \Theta$. For a target action $d_A$ in the interior of $\mathcal{D}$, strict concavity makes $d_A$ the unique Bayes action under $\pi_w$ if and only if it satisfies the first-order stationarity condition $\nabla_d \mathbb{E}_{\theta \sim \pi_w}[u(d, \theta)] \big|_{d=d_A} = \mathbf{0}$. Under regularity conditions permitting the interchange of expectation and differentiation, this is equivalent to $\mathbb{E}_{\theta \sim \pi_w}[\nabla_d u(d_A, \theta)] = \mathbf{0}$, so inducing a continuous decision again reduces to a finite-dimensional moment-matching problem, now with feature map given by the utility gradient at the target action, $\phi(\theta) = \nabla_d u(d_A, \theta)$. To meet the conditions of Theorem \ref{thm:obj_grad}, we assume that $\|\nabla_d u(d_A, \theta)\|_2$ has a finite second moment under both $\pi_w$ and $\pi_A$. Since optimality is now an equality, the target is $m = \mathbf{0}$, and the squared MMD becomes the squared norm of the expected utility gradient, $\mathrm{MMD}^2(\pi_A, \pi_w; k_{\mathrm{dec}}) = \| \mathbb{E}_{\theta \sim \pi_w}[\phi(\theta)] \|_2^2$. An attack that drives this objective to zero makes the gradient of the expected utility at $d_A$ vanish; since the expected utility inherits strict concavity from $u$, the resulting stationary point is the unique global maximizer, and hence $d^\star(w) = d_A$. Note that, if the objective is only driven below $\varepsilon^2$, $\kappa$-strong concavity of the expected utility still places the Bayes action within distance $\varepsilon/\kappa$ of $d_A$.

When the expected utility is not concave in $d$, the first-order condition is only necessary, so driving the objective to zero makes $d_A$ a stationary point but not necessarily the Bayes action. A practical remedy is to discretize: replace $\mathcal{D}$ by a finite grid containing $d_A$ and apply the finite-action construction of Section~\ref{sec:targeting_decisions}, which makes $d_A$ optimal over the grid. If the expected utility is $\lambda$-Lipschitz in $d$ and every action lies within distance $\tau$ of the grid, driving every expected gap below $-\lambda\tau$ further places the Bayes action within distance $\tau$ of $d_A$; since each grid point adds one feature, this suits low-dimensional actions.

\section{Additional Algorithmic Details}\label{app:algorithms}

For completeness, Algorithm \ref{alg:iscd} details the Integer-Steps Coordinate Descent (ISCD) heuristic introduced in Section \ref{sec:solution_methods}.

\begin{algorithm}[H]
\caption{Integer-Steps Coordinate Descent (ISCD)}\label{alg:iscd}
\begin{algorithmic}[1]
\State \textbf{Input:} Maximum number of iterations $T_{\text{max}}$, sample sizes $M$ and $N$; set $w \gets \mathbf{1}$ and $t = 0$.
\Repeat
    \State Sample $S_A \sim \pi_A$ (size $M$) and $S_{\pi_w} \sim \pi_w$ (size $2N$).
    \State Estimate the gradient $\hat{\nabla}$ with Algorithm~\ref{alg:grad_est} from $S_A$ and $S_{\pi_w}$.
    \State Choose the best feasible coordinate move:
    \[
       j \gets \argmax_{i \in \{1,\dots,n\}} \left\{|\hat{\nabla}_i| \ \Big| \ w - \text{sign}(\hat{\nabla}_i)e_i \in \mathcal{W} \right\}
    \]
    \State Update $w \gets w - \operatorname{sign}(\hat{\nabla}_j) e_j$ and $t \gets t + 1$.
\Until{the budget is exhausted, $\|w - \mathbf{1}\|_1 = B$, or $t \ge T_{\text{max}}$}
\State \Return $w$
\end{algorithmic}
\end{algorithm}

\clearpage

\section{Details of the Linear-Regression Experiments}

\subsection{Weighted posterior under the Normal-Inverse-Gamma prior}
\label{app:nig}

This appendix provides the derivation of the weighted posterior parameters for Bayesian linear regression with a Normal-Inverse-Gamma prior based on a perturbed dataset. The perturbation is represented by the weight vector $w \in \mathbb{R}^n$, or equivalently, the diagonal matrix $W = \text{diag}(w) \in \mathbb{R}^{n \times n}$. We parameterize the prior by its precision matrix $\Lambda_0 = \Sigma_0^{-1}$, where $\Sigma_0$ is the prior covariance scale used in Section~\ref{sec:linreg}; the clean-posterior scale matrix $\Sigma_n$ of that section is likewise $\Sigma_n = \Lambda_n^{-1}$.

The probabilistic model we are assuming is
\begin{equation*}
    \begin{aligned}
    \sigma^2 &\sim \text{Inv-Gamma}(a_0, b_0), \\
    \beta \,|\, \sigma^2 &\sim \mathcal{N}(\mu_0, \sigma^2 \Lambda_0^{-1}), \\
    y_i \,|\, x_i, \beta, \sigma^2 &\sim \mathcal{N}( x_i^\top \beta, \sigma^2),
\end{aligned}
\end{equation*}

where $\sigma^2 \sim \text{Inv-Gamma}(a_0, b_0)$ is equivalent to $\frac{1}{\sigma^2} \sim \text{Gamma}(a_0, b_0)$.
The pdf of $\sigma^2=s$ is given by
\begin{equation*}
    \quad f(s)=\frac{b_0^{a_0}}{\Gamma(a_0)} s^{-a_0-1} e^{-\frac{b_0}{s}}, \ \forall s> 0.
\end{equation*}

The joint NIG prior over the parameters is equal to
\begin{align*}
\pi\left(\beta, \sigma^{2}\right)
&= \text{Inv-Gamma}\left(\sigma^{2} \mid a_{0}, b_{0}\right) \cdot \mathcal{N}\left(\beta \mid \mu_{0}, \sigma^{2} \Lambda_0^{-1}\right) \\
&= \frac{b_0^{a_0}}{\Gamma\left(a_{0}\right)}
\left(\frac{1}{\sigma^{2}}\right)^{a_{0}+1} e^{-\frac{b_{0}}{\sigma^{2}}}
\frac{\left| \Lambda_0\right|^{\frac{1}{2}}}{(2 \pi \sigma^{2})^{\frac{d}{2}}} \exp\left(-\frac{1}{2 \sigma^{2}} (\beta - \mu_{0})^{\top}\Lambda_0(\beta - \mu_{0})\right).
\end{align*}

If we define $f_{X,y}(\beta, \sigma^{2})$ as the vector of log-likelihood contributions for each observation in the unperturbed dataset, then the likelihood of the weighted data, given the parameters $\beta$ and $\sigma^{2}$, can be expressed as:
\begin{align*}
\exp\left(w^{\top} f_{X,y}(\beta, \sigma^{2})\right)
&= \prod_{i=1}^n \left(\frac{1}{\sqrt{2\pi\sigma^2}}\right)^{w_i} \exp\left(-\frac{(y_i - x_i^\top \beta)^2}{2\sigma^{2}}\right)^{w_i}\\
&= \left(\frac{1}{\sqrt{2\pi\sigma^2}}\right)^{w^{\top}\mathbf{1}} \exp\left(-\frac{1}{2\sigma^{2}} \sum_{i=1}^n w_i(y_i - x_i^\top \beta)^2\right)\\
&= \frac{1}{\left(2\pi\sigma^2\right)^{\frac{w^{\top}\mathbf{1}}{2}} }
\exp\left(-\frac{1}{2\sigma^{2}}  (y - X \beta)^{\top} W (y - X \beta)\right).\\
\end{align*}

Thus, the posterior is proportional to
\begin{align*}
\pi_w\left(\beta, \sigma^{2} \mid X, y\right)
& \propto \; \pi\left(\beta, \sigma^{2}\right) \cdot \exp\left(w^{\top} f_{X,y}(\beta, \sigma^{2})\right) \\
& \propto\left(\frac{1}{\sigma^{2}}\right)^{a_{0}+1 + {\frac{d + w^{\top}\mathbf{1}}{2}}}
\exp\left(-\frac{b_{0}}{\sigma^{2}} -\frac{1}{2\sigma^{2}} Q(\beta)\right),
\end{align*}

where $Q(\beta)$ is defined as:
\begin{align*}\label{eq:multivariate_completion_square}
Q(\beta) &= (\beta - \mu_{0})^{\top}\Lambda_{0}(\beta - \mu_{0}) + (y - X\beta)^{\top}W(y - X\beta)\\
  &= \beta^{\top} \left(\Lambda_{0} + X^{\top}WX\right) \beta - 2\beta^{\top}\left(\Lambda_{0}\mu_{0} + X^{\top}Wy\right)  + \mu_{0}^{\top} \Lambda_{0}\mu_{0} + y^{\top}Wy  \\
  &= \beta^{\top}\Lambda_n\beta - 2\beta^{\top} \Lambda_n\mu_n + c\\
  &= (\beta - \mu_n)^{\top}\Lambda_n(\beta - \mu_n) - \mu_n^{\top}\Lambda_n\mu_n +c \\
  &= (\beta - \mu_n)^{\top}\Lambda_n(\beta - \mu_n) +c^{\ast},
\end{align*}

with
\begin{align*}
\Lambda_n &= \Lambda_0 + X^{\top}WX, \\
\mu_n &= \Lambda_n^{-1}\left(\Lambda_0\mu_{0} + X^{\top}Wy \right),\\
c &= \mu_{0}^{\top} \Lambda_0\mu_{0} + y^{\top}Wy, \\
c^{\ast} &= \mu_{0}^{\top} \Lambda_0 \mu_{0} + y^{\top}Wy - \mu_n^{\top}\Lambda_n\mu_n.
\end{align*}

Finally, we can express the posterior as a NIG distribution with parameters $(\mu_n, \Lambda_n, a_n, b_n)$, i.e.,

\begin{align*}
\pi_w\left(\beta, \sigma^{2} \mid X, y\right)
& \propto \left(\frac{1}{\sigma^{2}}\right)^{a_{0}+1 + \frac{d + w^{\top}\mathbf{1}}{2}}
\exp\left(-\frac{b_{0}}{\sigma^{2}}-\frac{1}{2 \sigma^{2}} ((\beta - \mu_n)^{\top}\Lambda_n(\beta - \mu_n) +c^{\ast})\right)\\
& \propto \left(\frac{1}{\sigma^{2}}\right)^{a_n+1} e^{-\frac{b_n}{\sigma^{2}}}
\left(\frac{1}{\sigma^2}\right)^{\frac{d}{2}}
\exp\left(-\frac{1}{2 \sigma^{2}} (\beta - \mu_n)^{\top}\Lambda_n(\beta - \mu_n)\right)\\
&= \text{Inv-Gamma}\left(\sigma^{2} \mid a_n, b_n \right) \cdot \mathcal{N}\left(\beta \mid \mu_n, \sigma^{2} \Lambda_n^{-1}\right), \\
\end{align*}

where
\begin{align*}
a_n &= a_{0} + \frac{1}{2}w^{\top}\mathbf{1}, \\
b_n &= b_{0} + \frac{c^{\ast}}{2}= b_{0}+\frac{1}{2}\left(\mu_{0}^{\top} \Lambda_{0} \mu_{0}+y^{\top} W y-\mu_n^{\top} \Lambda_n \mu_n\right)\;.
\end{align*}

\subsection{Implementation details}
\label{app:settings}

The experiments of Section~\ref{sec:linreg} and Appendix~\ref{exp:attack_var} share the following settings. The two covariates have unit variances, and Figure~\ref{fig:linreg_unknown_var_rho_sensitivity} uses one simulated dataset per $\rho$, so its $\rho=0.8$ panel differs slightly from Figure~\ref{fig:linreg_unknown_var_budget}. Each gradient estimate (Algorithm~\ref{alg:grad_est}) uses $2N=2000$ draws from $\pi_w$ and, for the RBF kernel, $M=1000$ draws from $\pi_A$. SGD-R2 and Adam-R2 start from $w_0=\mathbf{1}$ and run for at most $T=500$ iterations, stopping earlier once the first-order predicted change $|\hat{\nabla}_t^\top(w_{t+1}-w_t)|$ falls below $10^{-4}$ times its value at the first iteration. SGD-R2 takes normalized steps, $\eta_t = 0.5/\|\hat{\nabla}_t\|_2$, and Adam-R2 uses a constant learning rate $0.1$ with decay rates $(0.9, 0.999)$. ISCD runs for at most $T_{\max}=500$ iterations. The RBF kernel acts on $\theta=(\beta,\sigma^2)$, and its bandwidth is the median pairwise Euclidean distance in a pilot sample of $1000$ draws from each of the clean and adversarial posteriors. The reported MMD$^2$ is estimated from $2000$ draws of the attacked posterior (and, for the RBF kernel, $1000$ target draws through the unbiased U-statistic), whereas posterior means and variances are computed in closed form.

\subsection{Gradient unbiasedness check}\label{app:grad_check}

Lemma~\ref{lemma:grad_unbiased} shows that the gradient estimator of Algorithm~\ref{alg:grad_est} is unbiased. We check this numerically on the mean attack of Section~\ref{sec:linreg}, whose moment-matching objective $F(w)=(\mathbb{E}_{\theta \sim \pi_w}[\beta_1]-m_A)^2$, with target mean $m_A$, has the closed-form gradient
\begin{equation}
\nabla_w F(w) = 2\big(\mathbb{E}_{\theta \sim \pi_w}[\beta_1]-m_A\big)\,\big(y - X\mu_n\big) \odot \big(X\big[\Lambda_n^{-1}\big]_{:,1}\big),
\end{equation}
where $\mu_n$ and $\Lambda_n$ are as in Appendix~\ref{app:nig}, $[\Lambda_n^{-1}]_{:,1}$ is the first column of $\Lambda_n^{-1}$, and $\odot$ is the elementwise product. At $w=\mathbf{1}$, we compute $n_{\mathrm{rep}}=20000$ independent estimates $\hat{\nabla}$, each from $2N=100$ posterior draws, and compare their average $\bar{\nabla}$ with $\nabla F$ through the standardized errors
\begin{equation}
z_i = \frac{\bar{\nabla}_i - (\nabla F)_i}{\mathrm{se}_i},
\qquad
\mathrm{se}_i = \frac{\mathrm{sd}(\hat{\nabla}_i)}{\sqrt{n_{\mathrm{rep}}}},
\end{equation}
which are of order one if the estimator is unbiased.

\begin{figure}[t]
      \centering
      \includegraphics[width=\textwidth]{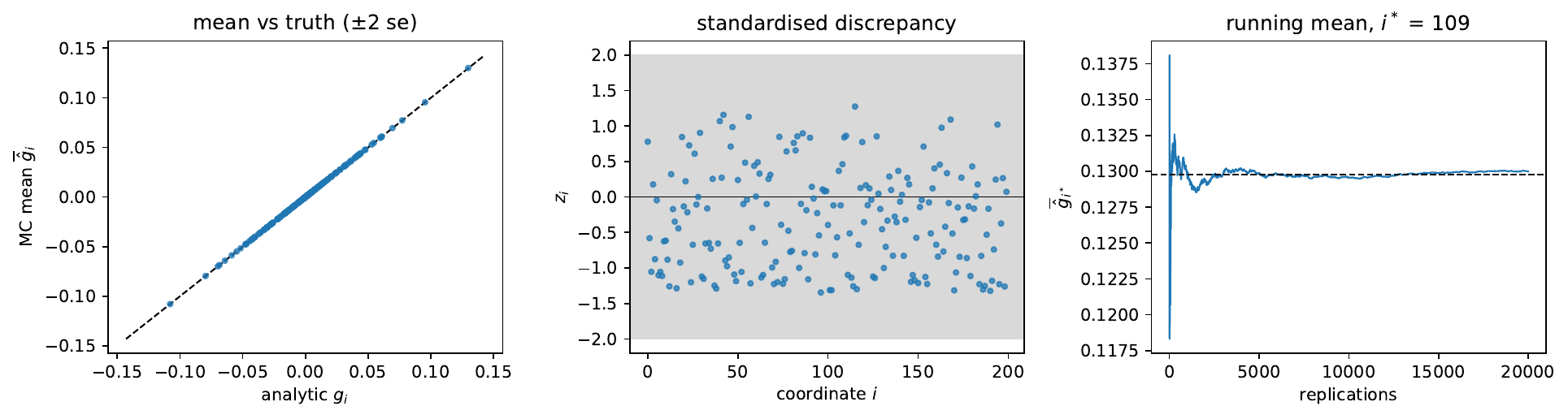}
      \caption{Unbiasedness of the stochastic MMD$^2$ gradient estimator at $w=\mathbf{1}$, over $n_{\mathrm{rep}}=20000$ replications. Left: Monte Carlo average of the estimator against the exact
  gradient, for each coordinate ($\pm 2$ standard errors). Middle: standardized discrepancy $z_i$; the shaded band is $|z_i|<2$. Right: running mean of the estimator at the coordinate
  with the largest absolute gradient; the dashed line is the exact value.}
      \label{fig:grad_unbiased}
  \end{figure}

Figure~\ref{fig:grad_unbiased} illustrates the results. The left panel plots, for each of the $n=200$ coordinates, the average of the estimates against the exact gradient, and all points lie on the diagonal. The middle panel shows the standardized errors $z_i$, all within $\pm 2$, with $\max_i|z_i|=1.35$. The right panel follows the running mean of the estimates at the coordinate with the largest absolute gradient: a single estimate is very noisy, but the running mean converges to the exact value.

\clearpage

\section{Attacking the Posterior Variance}
\label{exp:attack_var}

We consider the same problem as in Section \ref{sec:linreg}, but instead of attacking the mean we attack the variance of the marginal posterior over $\beta_1$, which is Student-t. The goal is to inflate this variance, so the target value $\sigma^2_A$ only needs to be well above the clean variance; we use $\sigma^2_A=0.5$ in all our experiments.
The mean is kept at its clean value $\bar{\beta}_1 = \mathbb{E}_{\theta \sim \pi_{\mathbf{1}}}[\beta_1]$, where $\pi_{\mathbf{1}}$ denotes the clean posterior, so that the attack changes the spread of the posterior rather than its location.

The moment-matching kernel can encode this goal in two ways. Both use the feature $h_1(\theta)=\beta_1$ with target $\bar{\beta}_1$, which pins the mean, and they differ in the second feature: the raw second moment $h_2(\theta)=\beta_1^2$, with target $\sigma^2_A+\bar{\beta}_1^{\,2}$, or the centered second moment $h_2(\theta)=(\beta_1-\bar{\beta}_1)^2$, with target $\sigma^2_A$. Only the centered feature yields a useful attack. To see why, write $V_w=\mathbb{V}_{\theta \sim \pi_w}[\beta_1]$ for the posterior variance, $V_{\mathbf{1}}$ for its clean value and, as in Section~\ref{sec:mm_kernel}, $\Delta_1=\mathbb{E}_{\theta \sim \pi_w}[\beta_1]-\bar{\beta}_1$ for the error in the mean. Since $\mathbb{E}_{\theta \sim \pi_w}[\beta_1^2]=V_w+(\bar{\beta}_1+\Delta_1)^2$ and $\mathbb{E}_{\theta \sim \pi_w}[(\beta_1-\bar{\beta}_1)^2]=V_w+\Delta_1^2$, the objective $F(w)=\mathrm{MMD}^2(\pi_A,\pi_w;k_{\text{MM}})$ is, under each feature,
\begin{equation*}
F_{\mathrm{raw}} = \Delta_1^2 + \big(V_w-\sigma^2_A+2\bar{\beta}_1\Delta_1+\Delta_1^2\big)^2,
\qquad
F_{\mathrm{cen}} = \Delta_1^2 + \big(V_w-\sigma^2_A+\Delta_1^2\big)^2 .
\end{equation*}
Both objectives attain their minimum, zero, exactly when the variance reaches the target without moving the mean, that is, when $V_w=\sigma^2_A$ and $\Delta_1=0$. They differ only in the term $2\bar{\beta}_1\Delta_1$ of $F_{\mathrm{raw}}$: the raw second moment grows not only when the posterior widens but also when its mean moves away from zero, so $F_{\mathrm{raw}}$ can be lowered by shifting the mean instead of widening the posterior, whereas $F_{\mathrm{cen}}$ depends on the mean only through $\Delta_1^2$. The gradients at the clean weights, where $\Delta_1=0$, make the difference explicit:
\begin{equation*}
\nabla_w F_{\mathrm{raw}} = 2(V_{\mathbf{1}}-\sigma^2_A)\big(\nabla_w V_w+2\bar{\beta}_1\nabla_w \Delta_1\big),
\qquad
\nabla_w F_{\mathrm{cen}} = 2(V_{\mathbf{1}}-\sigma^2_A)\,\nabla_w V_w,
\end{equation*}
where $\nabla_w \Delta_1=\mathrm{Cov}_{\pi_w}(\beta_1, f_{\mathbf{X}}(\theta))$, by Lemma~\ref{lemma:grad_cov}, is the direction followed by the mean attack of Section~\ref{sec:linreg}. The gradient of $F_{\mathrm{cen}}$ points along $\nabla_w V_w$, whereas that of $F_{\mathrm{raw}}$ adds the mean component $2\bar{\beta}_1\nabla_w\Delta_1$, which dominates when the mean is far from zero or easier to move than the variance. This is the case in our experiment, where $\bar{\beta}_1\approx 2$ and $V_{\mathbf{1}}=0.014$. Keeping the variance at $V_{\mathbf{1}}$, a shift of the mean by $0.11$ (about one posterior standard deviation) lowers $F_{\mathrm{raw}}$ by $94\%$ and leaves $F_{\mathrm{cen}}$ essentially unchanged; at the clean weights, the mean component $2\bar{\beta}_1\nabla_w \Delta_1$ is about $190$ times larger in norm than $\nabla_w V_w$. The raw attack therefore shifts the mean in its first step, which already removes most of $F_{\mathrm{raw}}$. Lowering it further would require undoing that shift while widening the posterior, but the variance responds to reweighting far less than the mean, so the attack stalls. As a result, at every budget the attained variance stays at or below its clean value.
 The centered objective has no such shortcut, and we use it in what follows.

Unlike the moment-matching kernel the RBF kernel requires a fully specified adversarial posterior $\pi_A$ to sample from. We take $\pi_A$ to be a normal-inverse-gamma distribution of the same form as the clean posterior, and modify only what the attack targets. We keep $a_n$ and $b_n$ at their clean values, so the marginal of $\sigma^2$ is untouched, and we keep the clean mean vector $\mu_n$, so no coefficient is displaced. The variance of the marginal of $\beta_1$, read off the Student-t marginal of $\beta$, whose covariance matrix is $\frac{b_n}{a_n-1}\Lambda_n^{-1}$, is then set by changing a single entry of $\Lambda_n^{-1}$,

  \begin{equation}
  \big[\Lambda_A^{-1}\big]_{11} = \sigma^2_A \, \frac{a_n - 1}{b_n}, \qquad \big[\Lambda_A^{-1}\big]_{jk} = \big[\Lambda_n^{-1}\big]_{jk} \ \text{ otherwise},
  \end{equation}

  so that $\frac{b_n}{a_n-1}\big[\Lambda_A^{-1}\big]_{11} = \sigma^2_A$ exactly. Changing the scale matrix rather than $a_n$ and $b_n$ is what allows the attack to inflate the
  variance of one coefficient while leaving the noise variance and the remaining coefficients at their clean values. Draws are then obtained hierarchically, as for the posterior
  itself: $1/\sigma^2 \sim \mathrm{Gamma}(a_n, b_n)$ followed by $\beta \mid \sigma^2 \sim \mathcal{N}(\mu_n, \sigma^2 \Lambda_A^{-1})$. Two remarks. First, because the off-diagonal entries are left unchanged while $\big[\Lambda_A^{-1}\big]_{11}$ grows, the target also has a weaker posterior correlation between
  $\beta_1$ and $\beta_2$. Second, inflating a
  diagonal entry preserves positive definiteness, so $\Lambda_A^{-1}$ is a valid scale matrix whenever $\sigma^2_A$ exceeds the clean marginal variance; targeting a smaller variance
  would require checking this explicitly.

\begin{figure}[tbp]
      \centering
      \begin{subfigure}[t]{0.62\linewidth}
          \centering
          \includegraphics[width=\linewidth]{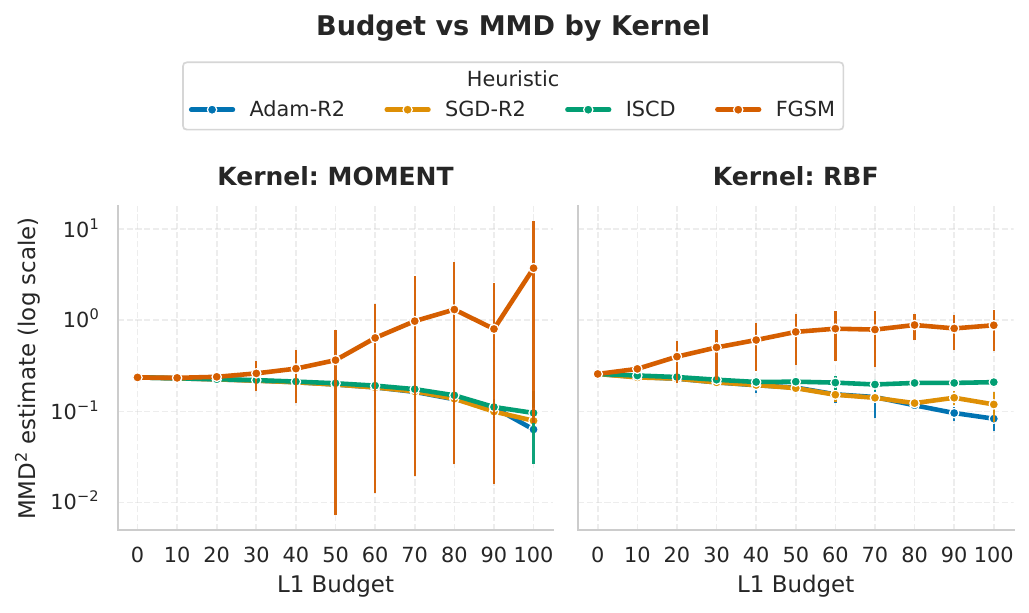}
          \caption{Achieved squared-MMD objective across heuristics, on a logarithmic scale.}
          \label{fig:linreg_unknown_var_variance_mmd_comparison}
      \end{subfigure}

      \medskip
      \begin{subfigure}[t]{\linewidth}
          \centering
          \includegraphics[width=\linewidth]{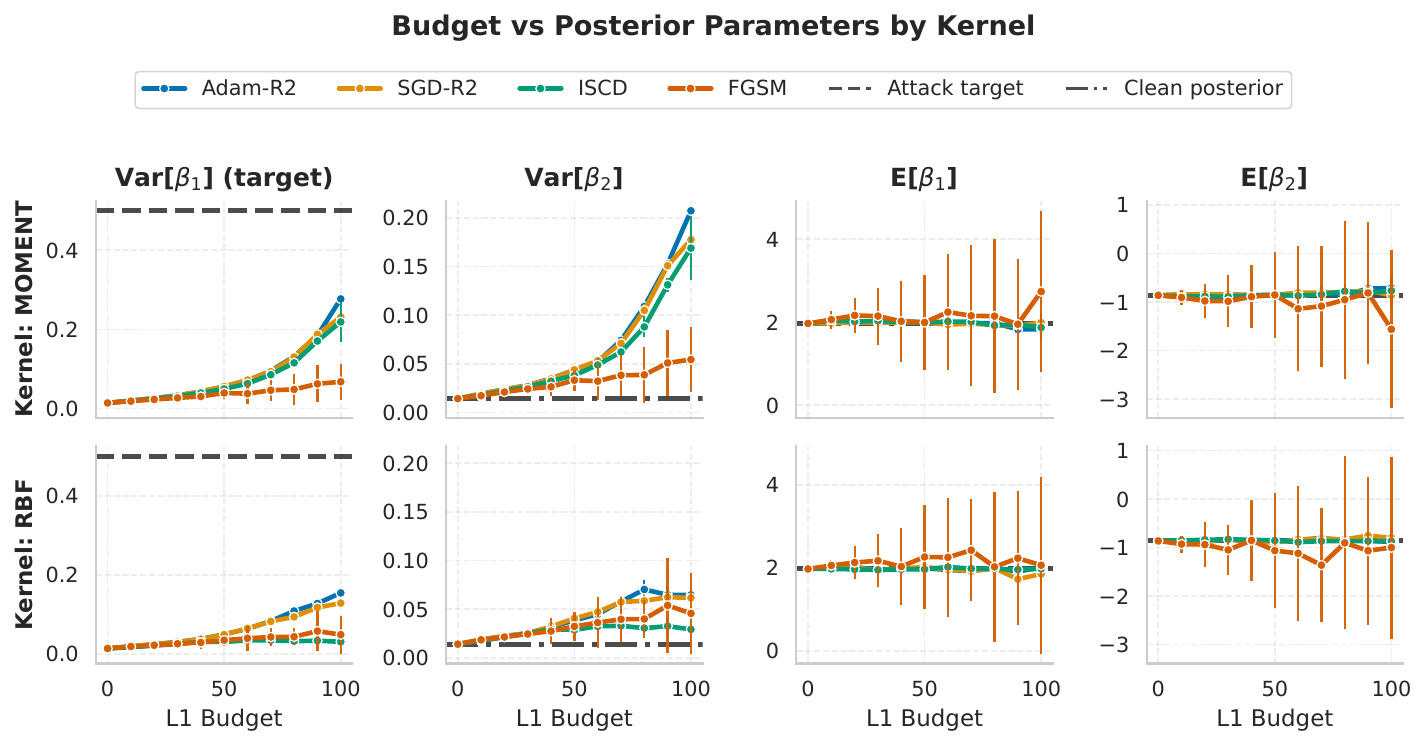}
          \caption{Induced posterior variances $\mathbb{V}_{\theta \sim \pi_w}[\beta_1]$ (targeted) and $\mathbb{V}_{\theta \sim \pi_w}[\beta_2]$ (non-targeted), and the posterior
  means $\mathbb{E}_{\theta \sim \pi_w}[\beta_1]$, $\mathbb{E}_{\theta \sim \pi_w}[\beta_2]$.}
          \label{fig:linreg_unknown_var_variance_posterior_parameters}
      \end{subfigure}
      \caption{Variance attack on the linear-regression posterior as a function of the perturbation budget $B$, by heuristic and kernel. The dashed line marks the adversarial
  target $\sigma^2_A$; dash-dotted lines mark the clean posterior values.}
      \label{fig:linreg_unknown_var_variance_budget}
  \end{figure}

\begin{figure}[tb]
      \centering
      \includegraphics[width=0.8\linewidth]{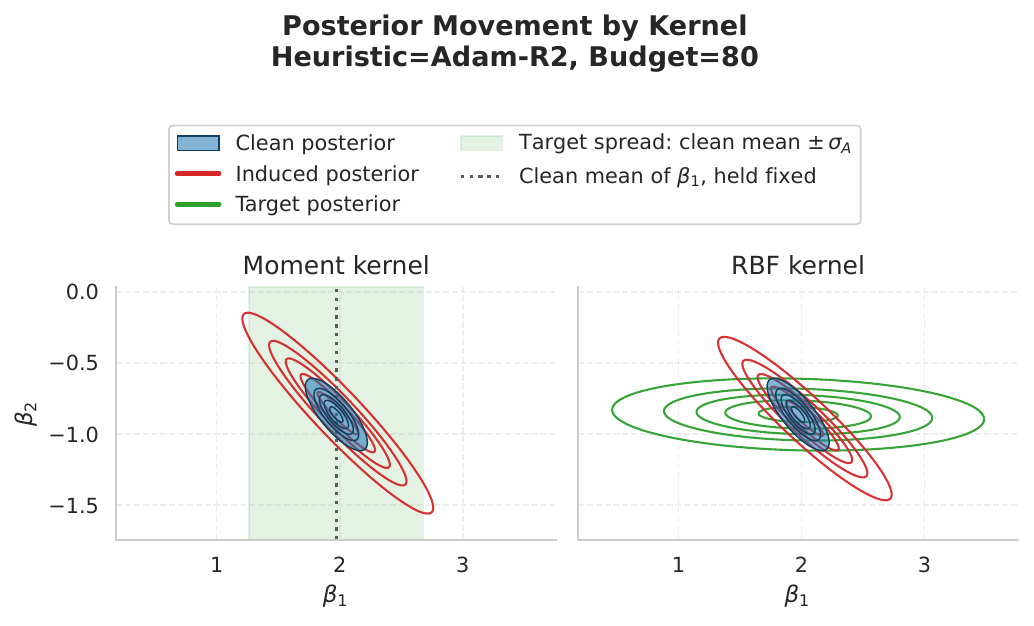}
      \caption{Posterior movement at budget $B=80$ for Adam-R2: clean, induced, and target posteriors of $(\beta_1,\beta_2)$ under the moment-matching (left) and RBF (right) kernels. In the moment-matching panel, the dotted line marks the clean mean $\bar{\beta}_1$, which the attack must preserve, and the shaded band spans $\bar{\beta}_1 \pm \sigma_A$, one target standard deviation on each side of it ($\sigma^2_A = 0.5$); no ellipse is drawn because the moment kernel does not define a full adversarial distribution. For each kernel, the run with the smallest MMD$^2$ is shown.}
      \label{fig:linreg_unknown_var_variance_posterior_movement}
  \end{figure}

Figure~\ref{fig:linreg_unknown_var_variance_budget} illustrates the variance attack as a function of the budget, for both kernels: panel~(a) shows the attained MMD$^2$ under each kernel, and panel~(b) the induced marginal posterior variances and means, with one row per kernel. Under the moment-matching kernel, the iterative heuristics inflate $\mathbb{V}_{\theta \sim \pi_w}[\beta_1]$ from $0.014$ to between $0.22$ and $0.28$ at $B=100$, while keeping $\mathbb{E}_{\theta \sim \pi_w}[\beta_1]$ within about $0.15$ of its clean value, as the first feature demands; FGSM, which commits its whole budget from a single gradient, inflates the variance far less and displaces the mean. The mean of $\beta_2$ also stays close to its clean value under the iterative heuristics, even though no feature constrains it.
As for the mean, the moment-matching kernel is more effective than the RBF kernel: with Adam-R2 it reaches
$\mathbb{V}_{\theta \sim \pi_w}[\beta_1] = 0.28$ against $0.15$ at the largest budget. Part of the gap is explained by what each kernel must specify. The moment-matching kernel constrains only the two features it lists, whereas the RBF kernel requires a fully specified $\pi_A$ to sample from, and every unconstrained feature of that distribution becomes an implicit target as well. In our construction $\pi_A$ inflates one entry of the scale matrix while leaving the off-diagonal untouched, which also drives the posterior correlation from $-0.77$ to $-0.13$. The RBF attack is therefore asked to decorrelate the posterior in addition to widening it, and reweighting cannot do both: the posterior correlation between $\beta_1$ and $\beta_2$ is essentially minus the correlation between the two predictors in the reweighted data, and making the predictors less correlated also shrinks the variance of $\beta_1$. At $B=100$ the induced correlation is $-0.89$ under Adam-R2, further from the target than the clean one.
Figure~\ref{fig:linreg_unknown_var_variance_posterior_movement} illustrates the geometry at $B=80$. Under both kernels, the induced posterior widens along the ridge of the clean posterior rather than along $\beta_1$. This is why, under the moment-matching kernel, the variance of $\beta_2$ grows almost as much as that of $\beta_1$ ($0.11$ against $0.13$ with Adam-R2), and why, under the RBF kernel, the induced contours never align with the target.

\begin{figure}[tb]
      \centering
      \includegraphics[width=\linewidth]{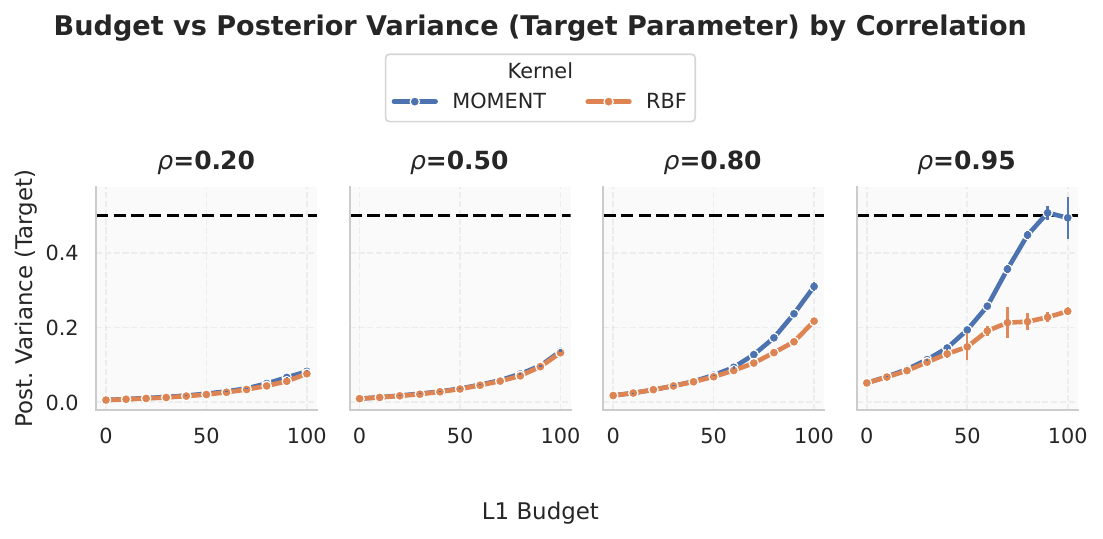}
      \caption{Variance attack under increasing predictor correlation. Each panel fixes the population correlation $\rho$ of the design and shows the attained $\mathbb{V}_{\theta \sim
  \pi_w}[\beta_1]$ against the budget, for both kernels. The adversarial target, marked by the dashed lines, is $\sigma^2_A = 0.5$ in every panel.}
      \label{fig:linreg_unknown_var_variance_rho}
  \end{figure}

Figure \ref{fig:linreg_unknown_var_variance_rho} illustrates how the attack depends on the population correlation $\rho$. Under the moment-matching kernel, the attained variance grows markedly with $\rho$, against the same target $\sigma^2_A=0.5$ in every panel: at $B=100$ it goes from $0.08$ at $\rho=0.2$ to $0.31$ at $\rho=0.8$, and at $\rho=0.95$ the attack reaches the target from $B=90$ on. The mechanism, however, is not that the attack itself becomes stronger: at a given budget, the multiplicative inflation it
  achieves is essentially the same at every correlation, around a factor of $9$ at $B=80$.
What changes is the starting point: with correlated predictors their effects are hard to tell apart, so the clean posterior of $\beta_1$ is wider to begin with, and its variance grows from $0.0057$ at $\rho=0.2$ to $0.0515$ at $\rho=0.95$. The same relative inflation therefore falls far short of the target at low correlation and reaches it at $\rho=0.95$.

\clearpage

\section{Diagnostics for the Spatial Attack}
\label{app:support_score}

\begin{figure}[htb]
    \centering
    \includegraphics[width=0.45\linewidth]{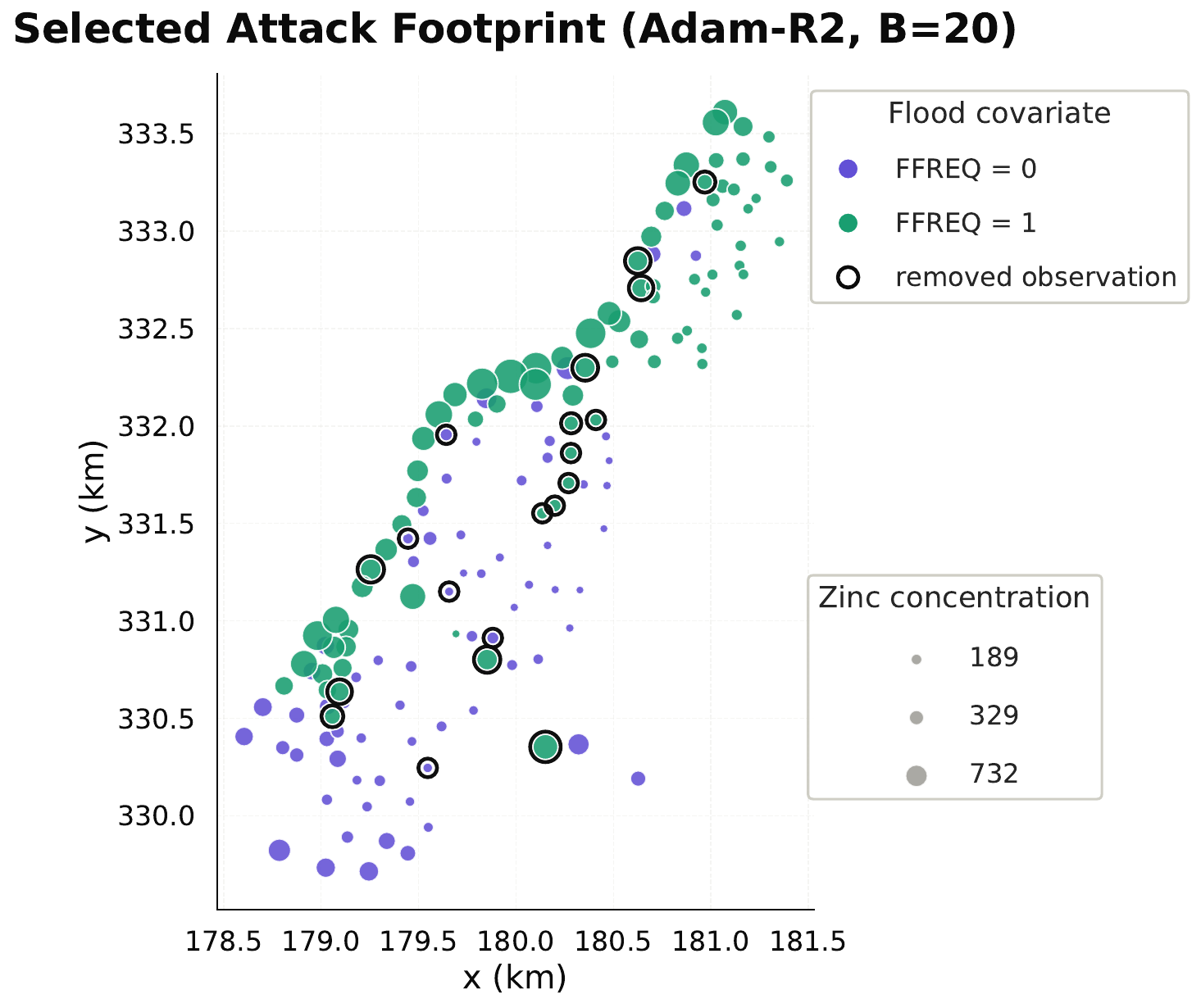}
    \caption{Footprint of the selected budget-$20$ Adam-R2 deletion attack. Marker size is proportional to zinc concentration; circled markers are removed observations.}
    \label{fig:spatial_attack_map}
\end{figure}

Figure~\ref{fig:spatial_attack_map} shows the footprint of the selected budget-$20$ Adam-R2 attack of Section~\ref{sec:spatial_regression}. The deleted sites are spread along the river corridor rather than concentrated in a single local cluster. The optimizer is not merely deleting the most obvious marginal outliers (e.g., exclusively selecting $\mathrm{FFREQ}=1$ sites with extremely high zinc concentrations, or $\mathrm{FFREQ}=0$ sites with extremely low concentrations). Instead, the deleted set includes a mix of both classes. This indicates that the attack is not relying on simple univariate extremes; rather, it is exploiting the complex interplay between the flooding indicator, the other covariates, and the latent spatial field.

To benchmark the optimized spatial attack of Section~\ref{sec:spatial_regression} against a naive influence-style heuristic, we define a simple proxy for how strongly a site \textit{supports} a positive flooding effect. Let $\theta$ denote a posterior draw of the parameters $(\beta, u, \sigma_\varepsilon, \sigma_u, \phi)$ of Section~\ref{sec:spatial_regression}.
For an observed $\mathrm{FFREQ}=1$ site, we compute the residual under the counterfactual mean obtained by setting $\mathrm{FFREQ}=0$ while keeping $\theta$ fixed; for an observed $\mathrm{FFREQ}=0$ site, we use the negative residual under the counterfactual $\mathrm{FFREQ}=1$ mean. Formally, if $\mu_i(f,\theta)$ denotes the fitted mean at site $i$ when the flood indicator is set to $f \in \{0,1\}$, the support score is
$$
r_i(\theta)
    =
    \begin{cases}
        y_i - \mu_i(0,\theta), & \text{if } \mathrm{FFREQ}_i = 1,\\[3pt]
        \mu_i(1,\theta) - y_i, & \text{if } \mathrm{FFREQ}_i = 0.
    \end{cases}
$$
Large positive values flag flooded sites whose zinc level is high relative to the no-flood counterfactual, and non-flooded sites whose zinc level is low relative to the flooded counterfactual; such sites pull $\beta_{\mathrm{FFREQ}}$ upward.

\begin{figure}[H]
    \centering
    \begin{subfigure}[t]{0.48\linewidth}
        \centering
        \includegraphics[width=\linewidth]{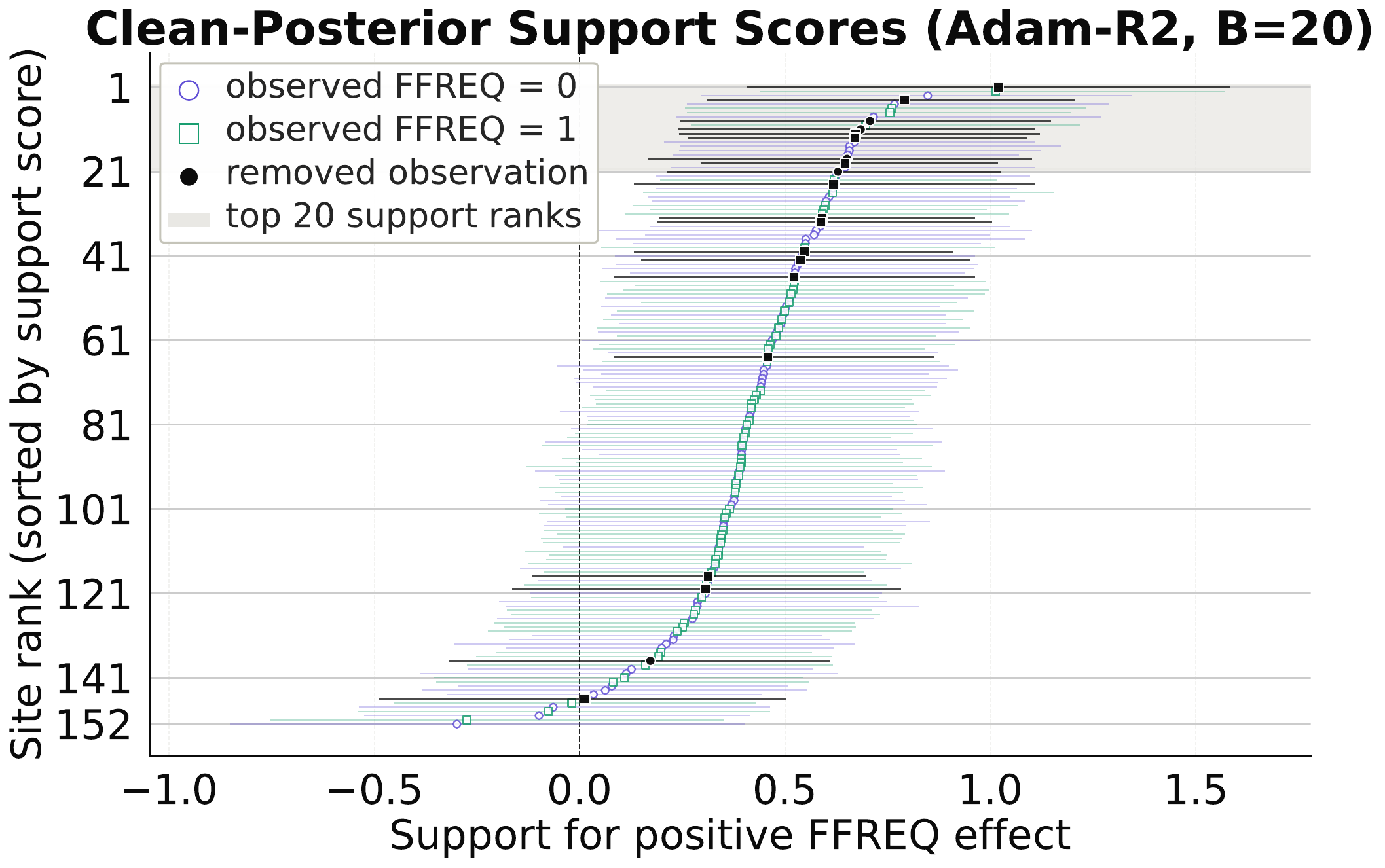}
        \caption{Clean-posterior support scores and the deleted observations.}
        \label{fig:spatial_support_scores}
    \end{subfigure}\hfill
    \begin{subfigure}[t]{0.48\linewidth}
        \centering
        \includegraphics[width=\linewidth]{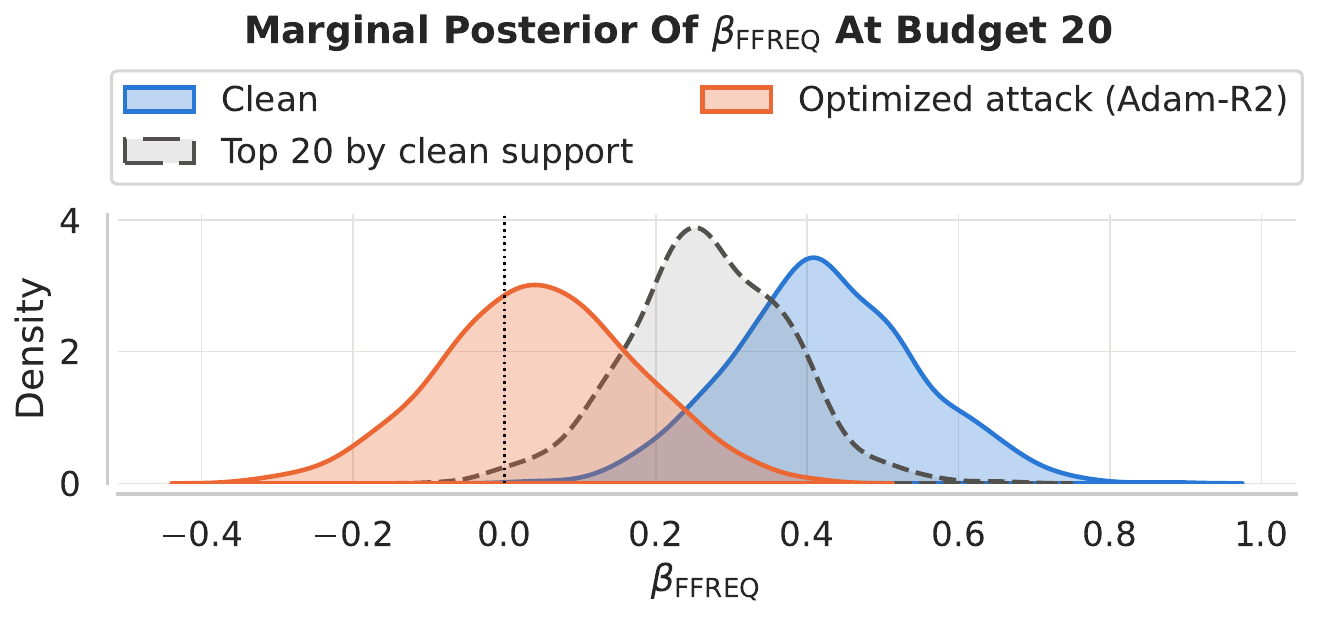}
        \caption{Posterior of $\beta_{\mathrm{FFREQ}}$: clean, after deleting the top-$20$ support-ranked sites, and under the optimized attack.}
        \label{fig:spatial_support_baseline}
    \end{subfigure}
    \caption{Support-score diagnostics for the selected budget-$20$ attack.}
    \label{fig:spatial_support_diagnostics}
\end{figure}

Given this intuition, one might assume that a successful attack could be executed simply by ranking the sites by their clean support scores and deleting the top $B$ of them. However, Figure~\ref{fig:spatial_support_diagnostics} shows that this heuristic is insufficient. Figure~\ref{fig:spatial_support_scores} displays the median and $95\%$ credible interval of $r_i(\theta)$ for each site under the clean posterior, with the sites ranked by this median. The sites deleted by the attack do tend to have higher-than-average support scores, but only $8$ of them belong to the top-$20$ support ranks. The limitations of a purely rank-based approach become explicit in Figure~\ref{fig:spatial_support_baseline}. Deleting the top $20$ sites ranked by clean support only lowers the posterior median of $\beta_{\mathrm{FFREQ}}$ to $0.27$, leaving a $95\%$ credible interval of $[0.05,\,0.48]$ that remains positive. By contrast, our optimized Adam-R2 attack drives the posterior median to $0.04$.

\section{Details of the Radon Decision Attack}\label{app:radon}

\emph{Expected health cost.} Write $m(\theta) = \alpha_{j_\star} + \beta_{\mathrm{floor}} x_\star$ and $\kappa = \log(C_{\max}/C_H)$. Since $\tilde y_\star \mid \theta \sim \mathcal{N}(m(\theta), \sigma_y^2)$, the conditional expected health cost of Section~\ref{sec:radon_decisions} is
$$
c(\theta) = C_H \exp\!\left(m(\theta) + \frac{\sigma_y^2}{2}\right) \Phi\!\left(\frac{\kappa - m(\theta) - \sigma_y^2}{\sigma_y}\right) + C_{\max}\left\{1 - \Phi\!\left(\frac{\kappa - m(\theta)}{\sigma_y}\right)\right\},
$$
where $\Phi$ is the standard normal distribution function.

\emph{Attack settings.} SGD-R2 starts from $w_0 = \mathbf{1}$ and runs for at most $24$ iterations with normalized steps of length $64/(t+1)$, estimating each gradient from $600$ draws of warm-started NUTS chains; Lake County homes cannot be deleted. With these settings, $B=7$ is the smallest budget that reverses the decision: with $B=6$, the posterior mean of the gap after the attack is $0.0$, so the decision is not reversed. Reported posterior expectations average four independent chains of $5000$ draws, with Monte Carlo standard errors of about $2$.

\begin{figure}[H]
    \centering
    \begin{subfigure}[t]{0.48\linewidth}
        \centering
        \includegraphics[width=\linewidth]{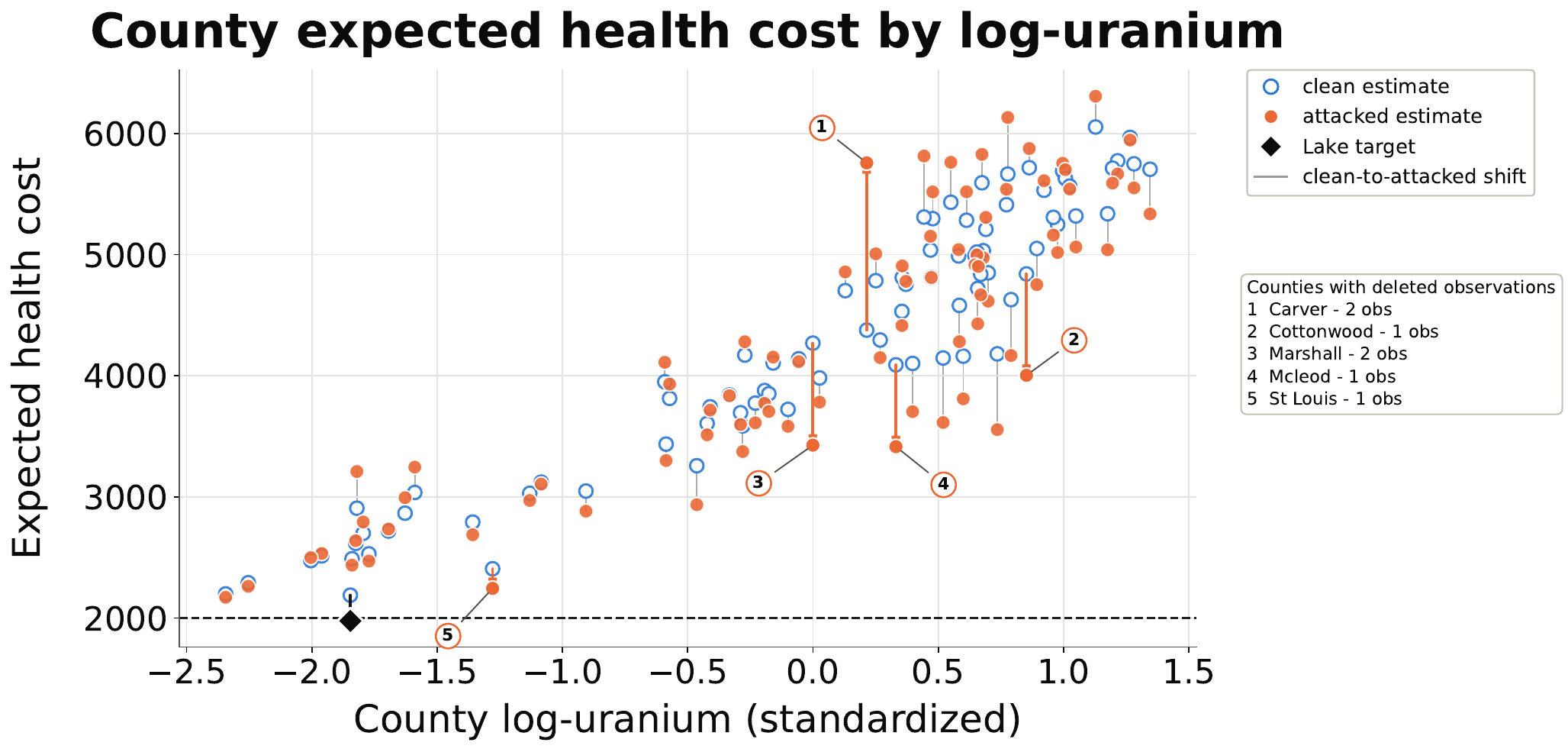}
        \caption{Expected health costs before and after the attack for a new home in each county, measured in the basement.}
        \label{fig:radon_county_cost}
    \end{subfigure}
    \hfill
    \begin{subfigure}[t]{0.48\linewidth}
        \centering
        \includegraphics[width=\linewidth]{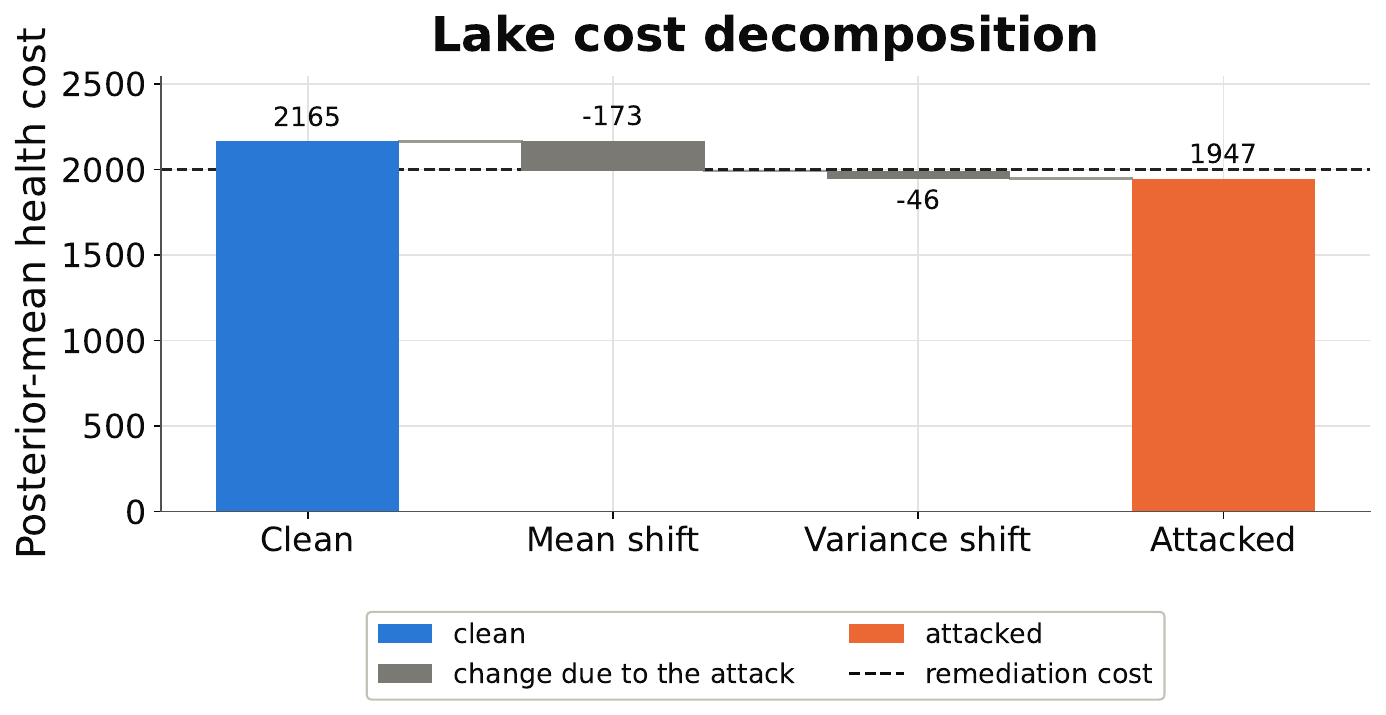}
        \caption{Decomposition of the Lake County expected health cost.}
        \label{fig:radon_cost_decomposition}
    \end{subfigure}

    \caption{Mechanism of the restricted radon attack.}
\label{fig:radon_county_mechanism}
\end{figure}

\emph{Mechanism.} The mechanism of the attack is fundamentally hierarchical. The seven deleted homes are spread across five non-Lake counties with different log-uranium levels: Carver (two observations), Cottonwood, Marshall (two observations), McLeod, and St.~Louis. Figure~\ref{fig:radon_county_cost} shows the expected health cost of a new basement home in every county under the clean and attacked data, highlighting the counties with deletions. Although no Lake County observation is deleted, the attack propagates through the partially pooled model and lowers Lake County's expected health cost, thereby reversing the Bayes action. Not all counties are affected in the same way: the expected costs fall in Lake, Cottonwood, Marshall, McLeod and St.~Louis counties, but rise in Carver County, whose deleted homes had low radon levels. For Lake County, Figure~\ref{fig:radon_cost_decomposition} shows a posterior-mean plug-in decomposition, which evaluates the expected health cost at the posterior-mean parameters and thereby separates the two channels through which the deletions act. Under the clean data this plug-in cost is about $2165$ (close to the posterior-mean cost of $2188$ reported in Section~\ref{sec:radon_decisions}) and sits above the $2000$ threshold. The attack lowers it in two ways: by reducing the predicted county intercept $\alpha_{j_\star}$, worth about $173$, and by shrinking the predictive variance $\sigma_y^2$, worth a further $46$, for an attacked plug-in cost of about $1947$. Because the utility gap equals the expected health cost minus $C_R$, both channels move the target household from the remediation side of the decision boundary to the no-action side.

\fi

\end{document}